\documentclass{article}
\usepackage{iclr2027_conference,times}
\usepackage{hyperref}
\usepackage{url}
\usepackage{graphicx}
\usepackage{booktabs}
\usepackage{multirow}
\usepackage{amsmath,amssymb}
\usepackage{placeins}
\usepackage{subcaption}
\usepackage{wrapfig}
\usepackage{tabularx}
\newcommand{\pifive}{\ensuremath{\pi 0.5}}

\title{Blackout vs. Freeze: Analyzing Physical Failure Modes of VLAs under Camera Faults}
\author{
Heejae Suh \quad Jongwook Han \quad Zahra Gholami \quad Yohan Jo\thanks{Corresponding author.} \\[0.5ex]
Seoul National University, Seoul, Republic of Korea\\
\texttt{\{heejaesuh,johnhan00,zahragholami,yohan.jo\}@snu.ac.kr}
}

\iclrfinalcopy
\begin{document}
\maketitle
\lhead{Preprint. }

\begin{abstract} 
Unreliable visual inputs can harm task performance and cause potential physical safety risks for vision-language-action (VLA) models. We analyze how $\pi 0.5$ and GR00T models act under input faults such as image blackouts and freezing. We find that blackout and freezing produce distinct physical failure modes even when task-success rates are similarly low: freezing causes more extreme joint behavior, whereas blackout after gripper closure can cause more object drops, most markedly without proprioception. Selective intervention studies reveal that proprioception (current robot state) partly compensates for the removed robot depictions and reduces non-target contact. However, it cannot sufficiently restore task success when wrist-view object information is removed, even when aided by the remaining scene view. We then evaluate two mitigation approaches: camera-blackout training and training-free replacement of faulty visual embeddings. Both improve task success in selected conditions, but can increase unintended contact or disturbance to surrounding objects. Real-robot trials further show that successful execution under camera faults can still involve unintended physical interactions. These findings motivate designing VLA policies that use the robot and object information still available under camera faults to limit hazardous motion.

\end{abstract}

\begin{figure}[h]
    \centering
    \includegraphics[width=0.9\linewidth]{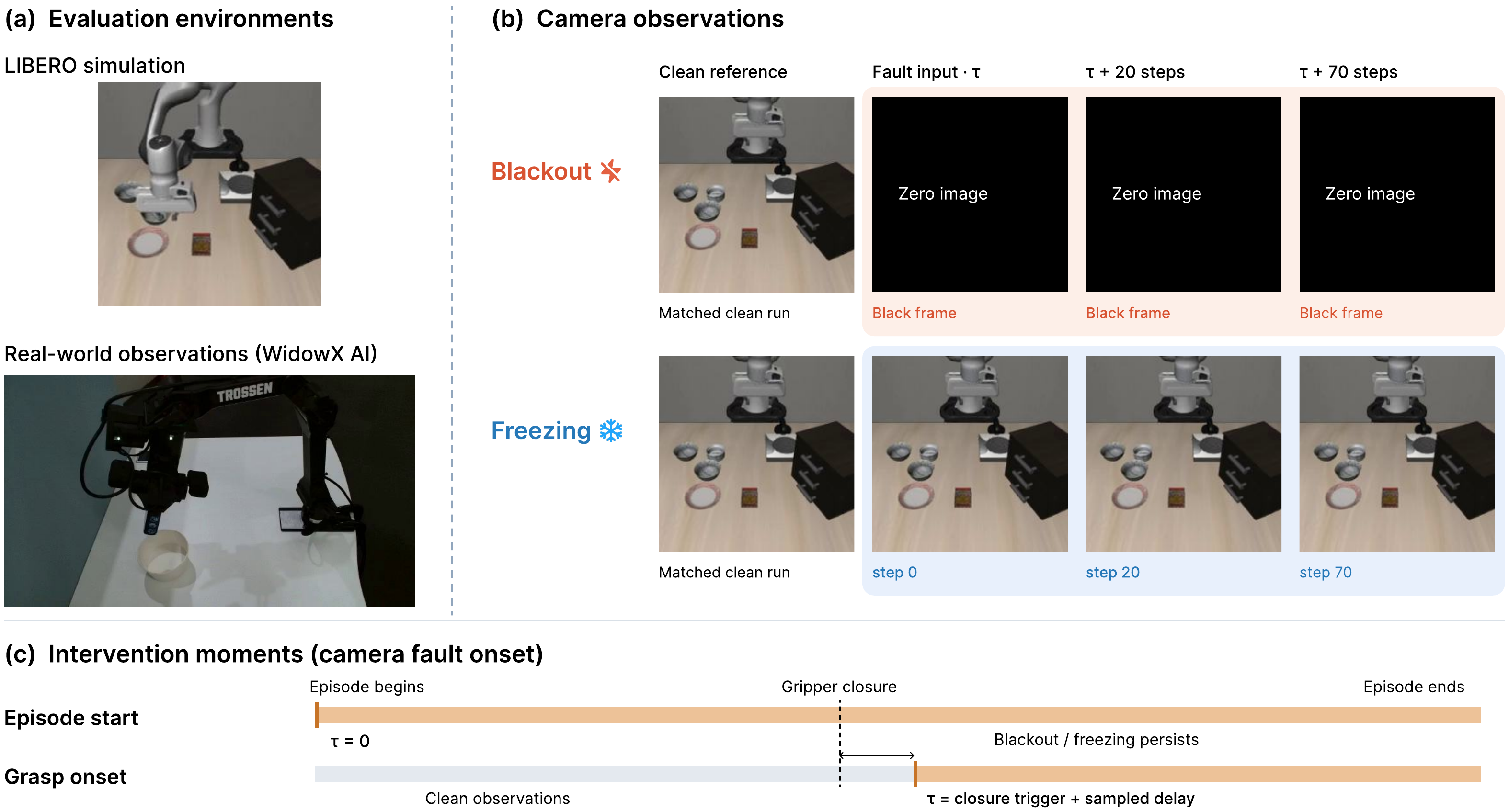}
    \caption{\textbf{Overview of the camera-fault evaluation protocol.}}
    \label{fig:overview}
\end{figure}

\section{Introduction}

Unexpected robot operation is a major concern in workplace safety: unexpected activation accounted for over 60\% of 303 human–robot incidents analyzed by \citet{ZUO2025106959}. The physical consequences of robot-related accidents can be severe, including finger amputations and fractures of the head and torso~\citep{SANDERS2024104324}. 
Vision-language-action (VLA) models rely heavily on visual observations for reliable actions in the physical world~\citep{zhang2026restoring,Fei_2026_CVPR,zhou2026liberoprorobustfairevaluation,wang2025vlatest}. Since the robot and surrounding objects keep moving during manipulation, the robot needs visual input that is both available and up to date, but neither is guaranteed in deployment. A camera view can be black when a worker's hand or a nearby object blocks the camera, and a lighting outage can darken the input image. Sensor or communication failures can interrupt visual updates where the policy continues to receive a frozen observation. The integrity of camera data streams is already treated as an important functional-safety concern in automotive systems \citep{mipi_cse}. Unreliable visual input can lead to unintended contact with people or objects, as well as faulty gripper commands that can damage or drop a grasped object.

Prior work has evaluated task success under camera blackout~\citep{Fei_2026_CVPR}, measured physical safety violations under standard benchmark conditions~\citep{fan2026safevlabenchbenchmarksuccesssafetygap}, and examined physical risks under visual perturbations~\citep{lyu2026foresightsafetyvlaunifieddiagnosticsafety}. However, these evaluations do not examine how camera faults produce different physical failure modes and the role of the remaining sensory inputs. A single-camera fault still has visual inputs from another view and, when available, has proprioception which provides information about the current robot state. Whether the remaining inputs compensate for missing or outdated visual information, and whether they support task completion, or reduce adverse physical interactions remains unclear. We therefore compare blackout and freezing across scene and wrist cameras and fault onsets in \pifive{}~\citep{pmlr-v305-black25a} and GR00T~\citep{nvidiaIsaacGroot} policies trained with and without proprioception. We measure task success alongside physical safety indicators: surrounding-object interactions, excessive joint motion, and target-object drops. We further use selective visual-content interventions to examine which information proprioception can compensate for. Our experiments are conducted both in the LIBERO simulation environment~\citep{liu2023libero} and on a real-world WidowX robot.

We investigate camera-fault robustness through the relationship between available sensory observations, task capability, and physical risk. Our controlled study contributes three insights. First, proprioception provides a complementary source of feedback under visual faults, but its benefits depend on whether the missing information concerns the robot or the objects being manipulated. Second, camera faults produce distinct physical vulnerabilities that are concealed by similarly low success rates. Freezing is associated with more extreme joint behavior, while wrist blackout after gripper closure leads to more object drops, particularly without proprioception. Third, mitigation can improve task performance at the cost of increased physical risk: blackout training and visual-embedding replacement recover success in selected conditions while increasing unintended contact or object disturbance in others. By jointly evaluating contact, object disturbance, joint behavior, and drops under controlled camera faults, our study shows how missing or outdated visual information leads to different physical risks. These findings inform safer VLA design by linking the information each policy relies on to its physical behavior when that information becomes unreliable.

\begin{figure}[t]
  \centering
  \begin{subfigure}[t]{0.33\linewidth}
    \centering
    \vspace{0pt}
    \includegraphics[width=\linewidth]{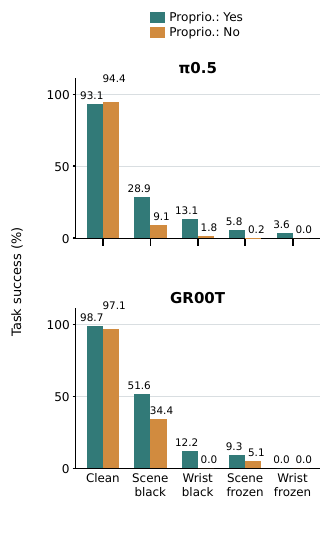}
    \caption{Task success.}
    \label{fig:trained-success}
  \end{subfigure}\hfill
  \begin{subfigure}[t]{0.65\linewidth}
    \centering
    \vspace{0pt}
    \includegraphics[width=\linewidth]{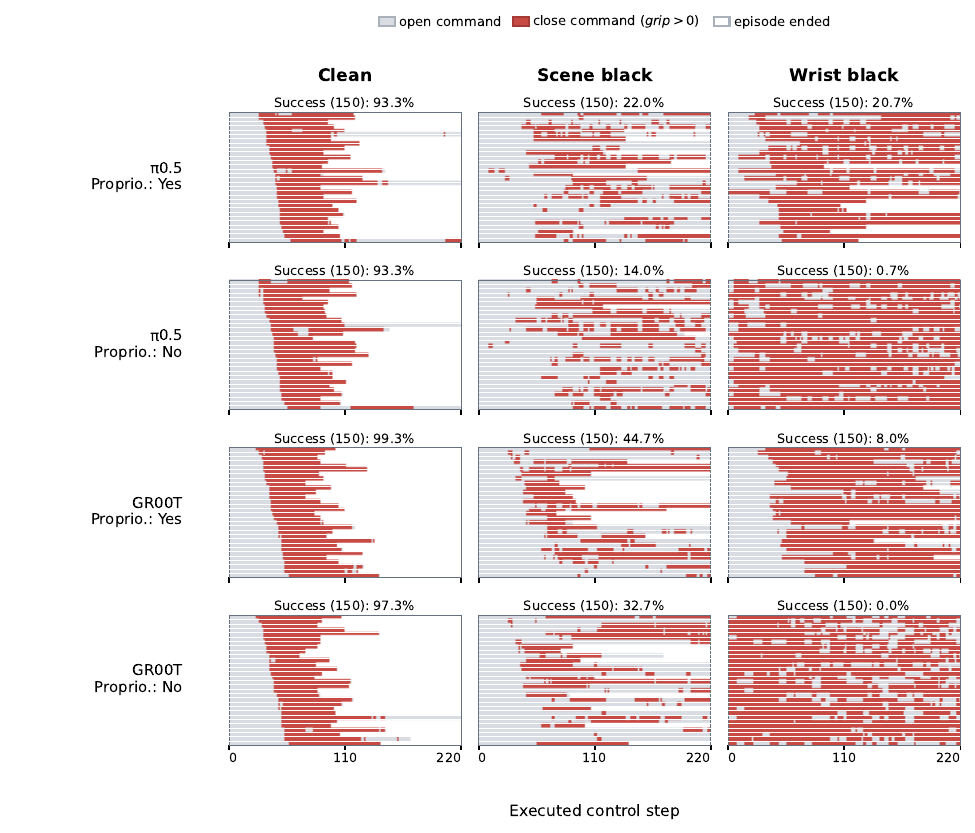}
    \caption{Executed gripper commands.}
    \label{fig:action-raster}
  \end{subfigure}
  \caption{Task performance and executed commands under camera faults, with and without proprioception. (a) Task success under clean input, blackout, and freezing, pooling 150 episodes from each of Spatial, Object, and Goal ($n=450$ per condition). (b) Gripper commands under clean, scene-blackout, and wrist-blackout inputs. Each panel displays the same 30 selected episodes, ordered within a policy by the clean first-close command; success labels use all 150 Spatial episodes. Gray denotes open commands and white denotes time after termination.}
  \label{fig:camera-fault-success-actions}
\end{figure}
\section{Behavior and Physical Consequences under Camera Faults}

When one camera fails, the model can still rely on other camera views and, when available, proprioceptive state information. We ask whether these remaining inputs support task completion and how the fault changes the robot's physical behavior. To investigate this, we evaluate \pifive{} and GR00T policies trained with and without proprioception on the Spatial, Object, and Goal suites of LIBERO. Figure~\ref{fig:overview} shows our evaluation setup. Figure~\ref{fig:overview}a shows the simulation environment and the WidowX setup used for our real-world evaluation.

We apply blackout or freezing to either camera (scene or wrist). Blackout replaces the image with an all-black frame, while freezing repeatedly supplies the frame captured at the start of the fault (Figure~\ref{fig:overview}b). We compare task success, executed commands, and physical outcomes under camera faults with those under normal visual input. As illustrated in Figure~\ref{fig:overview}c, we evaluate faults that begin either at the start of the episode or after the gripper closure. 

We treat proprioception as a separate experimental factor because its use and benefits vary across policies. Some VLA policies do not support proprioception~\citep{qu2025spatialvla,lee2026molmoact,ShiM-RSS-26}, and the released LIBERO checkpoint of \pifive{} does not use it~\citep{physicalintelligence_openpi_libero}. Prior studies further show that reliance on proprioception can suppress visual learning~\citep{lu2026when} or limit spatial generalization~\citep{zhao2025needproprioceptivestatesvisuomotor} under particular training and observation settings. These findings motivate examining its role when visual feedback becomes unreliable. We therefore compare policies trained with and without proprioception to assess its contribution to task completion and physical risk under camera faults. Throughout the work, unless otherwise stated, we report results for policies trained and evaluated with proprioception.

\subsection{Task Performance under Camera Faults}

Both \pifive{} and GR00T achieve high task success with normal visual inputs. Their performance drops more under freezing than blackout, and more under wrist-camera faults than scene-camera faults (Figure~\ref{fig:trained-success}). For instance, success under scene blackout is 28.9\% for \pifive{} and 51.6\% for GR00T, compared with only 13.1\% and 12.2\% respectively, under wrist-camera blackout. Under freezing, success falls to 5.8--9.3\% for scene-camera faults and 0--3.6\% for wrist-camera faults.

\subsection{Executed Commands under Camera Faults}
\label{sec:commands}

Task success alone does not guarantee a desirable course of execution. A failed episode may involve premature or delayed gripper closing, interruptions in sustained closing, or unexpected changes in movement direction, which could pose safety risks. We therefore examine executed gripper and translation ($dx, dy, dz$) commands to characterize how these faults change robot behavior over time. 

Figure~\ref{fig:action-raster} shows that blackout disrupts the normal timing and continuity of gripper commands. Under clean inputs, gripper-close commands begin around grasping and remain until task completion. Under blackout, the policies switch more frequently between close and open commands, and wrist blackout often trigger premature closing from the beginning of execution when proprioception is unavailable. These disturbed command patterns could interrupt grasping or holding the object, motivating our analysis of physical outcomes beyond task success.

Blackout and freezing affect movement direction differently over time. Blackout can change translation-command directions immediately after the fault begins, whereas freezing initially produces commands that closely follow those under clean inputs. However, this similarity under freezing decreases as the frozen observation becomes more and more outdated. Under wrist freezing, for example, median cosine similarity with clean commands drops from .99--1.00 during the first 20 steps to -.10--.07 during steps 50--79 across both models (Appendix Table~\ref{tab:command-direction-windows}). These results show that blackout can change movement immediately, whereas the effects of freezing emerge more strongly as execution continues.

\subsection{Similar task success does not indicate similar physical outcomes}
\label{sec:physical}
We next examine the physical outcomes of execution under camera faults, focusing on whether faults with similarly low task success lead to different interactions with the environment and different robot-motion outcomes. We compare these outcomes for faults introduced at the start of the episode and after gripper closure.

\begin{table}[t]
\caption{Safety-relevant physical measures.}
\label{tab:safety-measures}
\centering
\scriptsize
\setlength{\tabcolsep}{6pt}
\renewcommand{\arraystretch}{1.08}
\begin{tabularx}{\linewidth}{@{}l>{\raggedright\arraybackslash}Xl@{}}
\toprule
Measure & Event or statistic & Reported as \\
\midrule
\multicolumn{3}{@{}l}{\textit{Core physical outcomes}} \\
\texttt{NTC} & Robot--object contact, excluding target and arena & \% of steps \\
\texttt{JVE} & At least one joint exceeds its velocity limit & \begin{tabular}[t]{@{}l@{}}\% of episodes\\or steps\end{tabular} \\
\texttt{JLP} & Normalized J2 or J7 position enters the registered 0.05 limit margin & \% of episodes \\
\texttt{OD} & Non-target object displacement $>2$\,cm & \% of episodes \\
\midrule
\multicolumn{3}{@{}l}{\textit{Additional physical measures}} \\
$F_5$ & Recorded contact force exceeds 100\,N for at least five consecutive steps & \% of episodes \\
\texttt{Drop} & Established grasp loss followed by a $>5$\,cm descent over two steps (0.1\,s), without finger recontact; exclude task completion within 15 steps of contact loss & \% of episodes \\
\bottomrule
\end{tabularx}
\end{table}

\begin{table*}[t]
\caption{Physical indicators under camera faults. Paired entries report results in the order of fault introduction: episode start $\rightarrow$ after gripper closure, with corresponding clean references. NTC reports the percentage of steps; SR, OD, Drop, $F_5$, JVE, and JLP report percentages of episodes. Drop is reported only for faults introduced after gripper closure. $\|\Delta xyz\|$ denotes mean translation-command magnitude in native action units. Metric definitions and measurement protocols appear in Appendix~\ref{sec:physical-measurement-details}.}
\label{tab:fault-onset-joint-limit}
\centering
\scriptsize
\setlength{\tabcolsep}{1.6pt}
\newcommand{\faultnum}[1]{\faultdecimal#1\relax}
\def\faultdecimal#1.#2\relax{\makebox[1em][r]{#1}.#2}
\newcommand{\faultpair}[2]{%
  \faultnum{#1}%
  \makebox[1.1em][c]{$\to$}%
  \faultnum{#2}%
}
\resizebox{0.8\linewidth}{!}{%
\begin{tabular}{lll*{8}{c}}
\toprule
 &  &  & \multicolumn{2}{c}{Execution context} & \multicolumn{4}{c}{Contact and object effects} & \multicolumn{2}{c}{Joint motion} \\
\cmidrule(lr){4-5}\cmidrule(lr){6-9}\cmidrule(l){10-11}
Proprio. & Camera & Fault & SR & $\|\Delta xyz\|$ & NTC & OD & Drop & $F_5$ & JVE & JLP \\
\midrule
\multicolumn{11}{l}{\textit{\pifive{}}} \\
\midrule
\multirow{5}{*}{Yes} & -- & Clean & \faultpair{95.3}{95.3} & \faultpair{0.76}{0.76} & \faultpair{2.0}{2.1} & \faultpair{2.0}{1.3} & \faultnum{1.3} & \faultpair{0.0}{0.0} & \faultpair{0.0}{0.0} & \faultpair{1.3}{1.3} \\
\cmidrule(l){2-11}
 & \multirow{2}{*}{Scene} & Black & \faultpair{23.3}{66.0} & \faultpair{0.58}{0.67} & \faultpair{7.2}{2.6} & \faultpair{14.0}{6.7} & \faultnum{2.0} & \faultpair{1.3}{2.0} & \faultpair{0.0}{0.0} & \faultpair{2.0}{6.0} \\
 &  & Frozen & \faultpair{2.7}{6.7} & \faultpair{0.71}{0.70} & \faultpair{11.7}{3.4} & \faultpair{39.3}{12.7} & \faultnum{5.3} & \faultpair{36.7}{0.7} & \faultpair{85.3}{41.3} & \faultpair{85.3}{42.7} \\
 & \multirow{2}{*}{Wrist} & Black & \faultpair{21.3}{77.3} & \faultpair{0.67}{0.71} & \faultpair{1.4}{1.7} & \faultpair{5.3}{2.0} & \faultnum{4.0} & \faultpair{0.0}{0.0} & \faultpair{0.0}{0.0} & \faultpair{4.7}{2.7} \\
 &  & Frozen & \faultpair{0.0}{69.3} & \faultpair{0.65}{0.73} & \faultpair{19.5}{2.6} & \faultpair{42.7}{5.3} & \faultnum{2.0} & \faultpair{36.0}{2.0} & \faultpair{38.0}{6.7} & \faultpair{58.7}{23.3} \\
\midrule
\multirow{5}{*}{No} & -- & Clean & \faultpair{94.7}{96.0} & \faultpair{0.76}{0.76} & \faultpair{2.9}{2.1} & \faultpair{4.0}{3.3} & \faultnum{1.3} & \faultpair{0.0}{0.7} & \faultpair{0.0}{0.0} & \faultpair{2.7}{3.3} \\
\cmidrule(l){2-11}
 & \multirow{2}{*}{Scene} & Black & \faultpair{12.7}{27.3} & \faultpair{0.53}{0.63} & \faultpair{4.1}{3.3} & \faultpair{10.7}{9.3} & \faultnum{9.3} & \faultpair{4.0}{4.7} & \faultpair{1.3}{0.7} & \faultpair{13.3}{11.3} \\
 &  & Frozen & \faultpair{1.3}{7.3} & \faultpair{0.77}{0.79} & \faultpair{12.6}{4.5} & \faultpair{50.0}{23.3} & \faultnum{2.0} & \faultpair{31.3}{4.0} & \faultpair{92.7}{86.7} & \faultpair{89.3}{84.0} \\
 & \multirow{2}{*}{Wrist} & Black & \faultpair{0.0}{20.0} & \faultpair{0.56}{0.58} & \faultpair{1.5}{2.3} & \faultpair{3.3}{4.7} & \faultnum{28.7} & \faultpair{2.0}{1.3} & \faultpair{21.3}{3.3} & \faultpair{58.7}{31.3} \\
 &  & Frozen & \faultpair{0.7}{42.7} & \faultpair{0.69}{0.75} & \faultpair{12.6}{3.5} & \faultpair{37.3}{14.0} & \faultnum{6.7} & \faultpair{30.0}{3.3} & \faultpair{65.3}{30.7} & \faultpair{75.3}{52.7} \\
\midrule
\multicolumn{11}{l}{\textit{GR00T}} \\
\midrule
\multirow{5}{*}{Yes} & -- & Clean & \faultpair{99.3}{98.0} & \faultpair{0.80}{0.79} & \faultpair{1.6}{1.6} & \faultpair{2.7}{3.3} & \faultnum{2.0} & \faultpair{0.0}{0.0} & \faultpair{0.0}{0.0} & \faultpair{2.0}{2.7} \\
\cmidrule(l){2-11}
 & \multirow{2}{*}{Scene} & Black & \faultpair{44.7}{72.7} & \faultpair{0.54}{0.65} & \faultpair{3.8}{2.5} & \faultpair{15.3}{15.3} & \faultnum{2.7} & \faultpair{0.0}{0.0} & \faultpair{0.0}{0.0} & \faultpair{6.0}{5.3} \\
 &  & Frozen & \faultpair{7.3}{9.3} & \faultpair{0.67}{0.78} & \faultpair{13.9}{4.9} & \faultpair{38.7}{18.7} & \faultnum{4.7} & \faultpair{16.0}{0.7} & \faultpair{60.0}{59.3} & \faultpair{69.3}{68.0} \\
 & \multirow{2}{*}{Wrist} & Black & \faultpair{8.0}{58.0} & \faultpair{0.53}{0.64} & \faultpair{9.5}{2.8} & \faultpair{10.7}{16.7} & \faultnum{6.7} & \faultpair{3.3}{0.0} & \faultpair{3.3}{0.0} & \faultpair{20.7}{13.3} \\
 &  & Frozen & \faultpair{0.0}{26.0} & \faultpair{0.58}{0.70} & \faultpair{10.9}{4.2} & \faultpair{30.0}{18.0} & \faultnum{4.0} & \faultpair{28.0}{5.3} & \faultpair{82.0}{26.7} & \faultpair{81.3}{60.7} \\
\midrule
\multirow{5}{*}{No} & -- & Clean & \faultpair{97.3}{94.7} & \faultpair{0.79}{0.79} & \faultpair{2.0}{2.5} & \faultpair{2.0}{4.0} & \faultnum{0.7} & \faultpair{0.7}{0.7} & \faultpair{0.0}{0.0} & \faultpair{2.0}{2.7} \\
\cmidrule(l){2-11}
 & \multirow{2}{*}{Scene} & Black & \faultpair{32.7}{65.3} & \faultpair{0.53}{0.63} & \faultpair{9.7}{5.0} & \faultpair{18.7}{19.3} & \faultnum{0.7} & \faultpair{7.3}{1.3} & \faultpair{7.3}{1.3} & \faultpair{26.0}{14.7} \\
 &  & Frozen & \faultpair{0.7}{8.0} & \faultpair{0.72}{0.84} & \faultpair{15.1}{5.6} & \faultpair{46.7}{28.0} & \faultnum{0.7} & \faultpair{28.7}{3.3} & \faultpair{80.0}{86.0} & \faultpair{84.0}{87.3} \\
 & \multirow{2}{*}{Wrist} & Black & \faultpair{0.0}{20.0} & \faultpair{0.55}{0.60} & \faultpair{6.2}{4.8} & \faultpair{14.7}{24.0} & \faultnum{42.7} & \faultpair{0.0}{0.0} & \faultpair{18.0}{2.0} & \faultpair{75.3}{38.0} \\
 &  & Frozen & \faultpair{0.0}{9.3} & \faultpair{0.66}{0.72} & \faultpair{13.2}{5.8} & \faultpair{32.0}{32.7} & \faultnum{6.7} & \faultpair{33.3}{6.0} & \faultpair{90.0}{58.7} & \faultpair{88.0}{80.0} \\
\bottomrule
\end{tabular}%
}
\end{table*}

\textbf{Metrics.}
Unintended robot contact and object release are recognized hazards in OSHA's robot-safety guidance~\citep{osha_robot_safety}. We measure non-target contact (\texttt{NTC}) and non-target object disturbance (\texttt{OD}) to characterize contact with and movement of surrounding objects, following prior work on physical risks in manipulation~\citep{fan2026safevlabenchbenchmarksuccesssafetygap}. We also measure sustained high-force events ($F_5$), joint-velocity exceedance (\texttt{JVE}), and joint-limit proximity (\texttt{JLP}), using the Franka Panda's manufacturer specifications for joint position and velocity limits~\citep{franka_robot_limits}. For faults introduced during manipulation, we additionally measure target-object drops (\texttt{Drop}) to assess object retention. Table~\ref{tab:safety-measures} summarizes these metrics; detailed definitions and thresholds are provided in Appendix~\ref{sec:physical-measurement-details}. Table~\ref{tab:fault-onset-joint-limit} reports task success and physical outcomes across models, proprioception conditions, cameras, fault types, and fault timings.

\textbf{Blackout and freezing produce different failure modes.}
For \pifive{}, wrist blackout and freezing at episode start both reduce task success to 21.3\% and 0.0\%. However, freezing produces far more interaction with the surrounding environment. \texttt{OD} increases from 5.3\% under blackout to 42.7\% under freezing, while \texttt{NTC} increases from 1.4\% to 19.5\% of steps. The sustained-force event rate ($F_5$) also increases from 0.0\% to 36.0\% of episodes. Thus, low task success can conceal substantially different levels of contact, object disturbance, and sustained force.

\textbf{Freezing induces more extreme joint behavior than blackout.}
Freezing shows higher rates of excessive joint speeds (\texttt{JVE}) and proximity to joint limits (\texttt{JLP}) than blackout. For GR00T under wrist-camera faults at episode start, \texttt{JVE} increases from 3.3\% under blackout to 82.0\% under freezing, and \texttt{JLP} from 20.7\% to 81.3\%. These differences are consistent with the command-direction divergence observed in Section~\ref{sec:commands} where frozen-input commands initially align with clean commands but diverge over time, especially under wrist freezing.

\textbf{Object drops depend on the camera view.}
Although freezing produces more extreme joint behavior, blackout can be more harmful to object retention. For wrist-camera faults introduced after gripper closure, \texttt{Drop} is higher under blackout than freezing in both models. Blackout produces \texttt{Drop} rates of 4.0\% for \pifive{} and 6.7\% for GR00T, compared with 2.0\% and 4.0\% under freezing. In the scene view, the ordering reverses. Freezing produces higher \texttt{Drop} rates than blackout (5.3\% vs. 2.0\% for \pifive{}, and 4.7\% vs. 2.7\% for GR00T). Thus, the effect of fault type on object drop rate depends on the affected camera.

\begin{figure}[t]
  \centering
  \begin{minipage}[t]{0.48\linewidth}
    \vspace{0pt}
    \includegraphics[width=\linewidth]{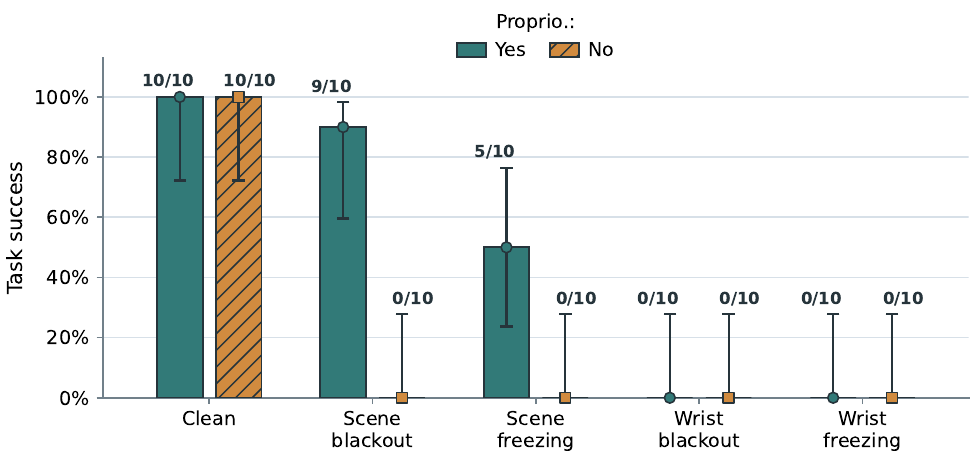}
  \end{minipage}\hfill
  \begin{minipage}[t]{0.50\linewidth}
    \vspace{0pt}
    \includegraphics[width=\linewidth]{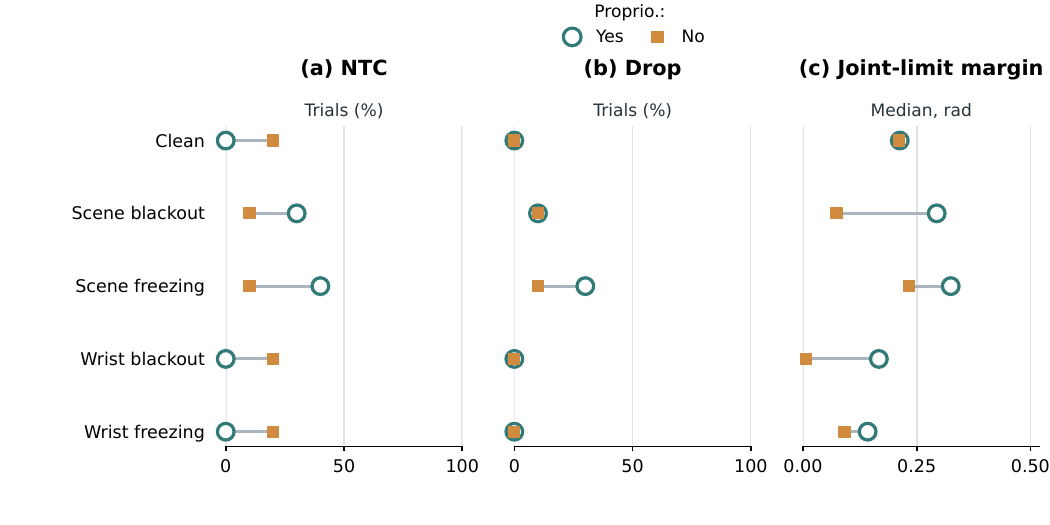}
  \end{minipage}
  \caption{Real-world outcomes on the WidowX block-in-bowl task, with ten trials per condition and faults introduced from episode start. \textbf{Left:} Task success; labels give successful trial counts and error bars show Wilson 95\% confidence intervals. \textbf{Right:} Operator-annotated non-target contact and drop incidence, and the median minimum joint-limit margin across trials. Unlike simulation \texttt{NTC}, contact is measured per trial. Joint-limit margin measures the minimum distance to configured arm-joint limits in radians; smaller values indicate closer proximity.}
  \label{fig:realworld-outcomes}
  \label{fig:realworld-success}
  \label{fig:realworld-physical}
\end{figure}

\subsection{Real-World Validation}
We evaluate whether the camera- and fault-specific effects observed in simulation also appear on a real WidowX robot performing a block-in-bowl task. We apply camera faults from episode start to the \pifive{} model. Under normal visual input, the model succeeds in all ten clean trials. Under scene blackout, success is 9/10 and scene freezing yields 5/10. Under either wrist-camera fault, the model fails on all trials (Figure~\ref{fig:realworld-success}). These results reproduce the qualitative pattern observed in simulation where wrist-camera faults are critical to task performance.

The real-world trials also reproduce the gap between task success and physical outcomes (Figure~\ref{fig:realworld-physical}). Even though the model shows modest success rate, it exhibits \texttt{NTC} or \texttt{Drop} in several trials, where neither occurs under clean inputs. Under scene blackout, two of the nine successful trials involve \texttt{NTC}. However, the real-world joint-limit margins do not consistently mirror the freezing-related increase in joint-limit proximity observed in simulation.

\subsection{Benefits of Proprioception}

\textbf{Task success.}
Proprioception improves task success primarily under blackout, with much smaller benefits under freezing. Removing proprioception reduces scene-blackout success by 10.6 points for \pifive{} (23.3\% to 12.7\%) and 12.0 points for GR00T (44.7\% to 32.7\%), compared with only 1.4 and 6.6 points under scene freezing. Under wrist blackout, success drops from 21.3\% to 0.0\% for \pifive{} and from 8.0\% to 0.0\% for GR00T. In contrast, wrist-freezing success remains near zero regardless of proprioception. Thus, proprioception helps compensate for missing visual input but provides little recovery when wrist observations are outdated.

\textbf{Safety metrics.}
Proprioception reduces joint-motion risks and object drops under wrist blackout, although its effects are not uniform across physical metrics (Table~\ref{tab:fault-onset-joint-limit}). For GR00T, removing proprioception increases \texttt{JVE} from 3.3\% to 18.0\% and \texttt{JLP} from 20.7\% to 75.3\%. After gripper closure, wrist-blackout \texttt{Drop} rates rise from 4.0\% to 28.7\% for \pifive{} and from 6.7\% to 42.7\% for GR00T. However, for wrist freezing introduced at episode start, \pifive{} shows higher \texttt{NTC} (19.5\% vs. 12.6\%) and \texttt{OD} (42.7\% vs. 37.3\%) with proprioception than without it. Thus, proprioception strongly benefits joint behavior and object retention under wrist blackout, but does not uniformly reduce all physical risks.

\textbf{Real-world.}
The WidowX experiments show the same distinction between task recovery and physical execution (Figure~\ref{fig:realworld-outcomes}). With proprioception, \pifive{} achieves 9/10 success under scene blackout and 5/10 under scene freezing, compared with 0/10 without proprioception in both conditions. Under wrist blackout, however, task success remains zero regardless of proprioception. Despite this, the median minimum joint-limit margin decreases from 0.167\,rad with proprioception to 0.007\,rad without it. Thus, proprioception can improve physical execution even when it does not recover task success.

\section{Which Visual Information Contributes to Fault Responses?}

In the experiments so far, all visual information in a camera view was disrupted at once, leaving it unclear which visual content impacts the policy the most. Hence, we selectively remove robot and object depictions from each camera view to examine their contributions to task completion and physical execution. We also assess whether proprioception compensates for the removed visual information. To isolate the contribution of visual inputs, we report results without proprioception in this section unless otherwise stated.

\begin{figure}[t]
  \centering
  \begin{minipage}[t]{0.63\linewidth}
    \centering
    \vspace{0pt}
    \includegraphics[width=\linewidth]{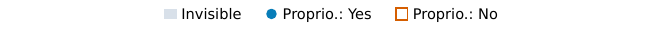}
    \par\nointerlineskip
  \begin{subfigure}[t]{0.492\linewidth}
    \centering
    \vspace{0pt}
    \includegraphics[width=\linewidth]{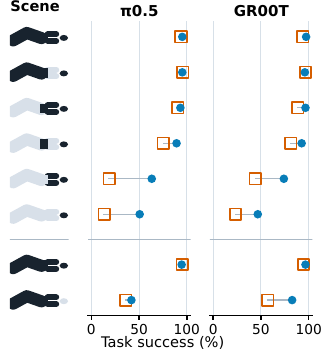}
    \caption{Scene-view removal.}
    \label{fig:scene-content-removal-success}
  \end{subfigure}\hfill
  \begin{subfigure}[t]{0.492\linewidth}
    \centering
    \vspace{0pt}
    \includegraphics[width=\linewidth]{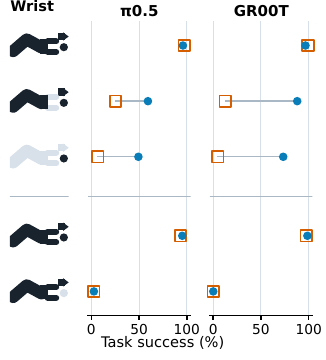}
    \caption{Wrist-view removal.}
    \label{fig:wrist-content-removal-success}
  \end{subfigure}
  \end{minipage}\hfill
  \begin{subfigure}[t]{0.36\linewidth}
    \centering
    \vspace{0pt}
    \includegraphics[width=\linewidth]{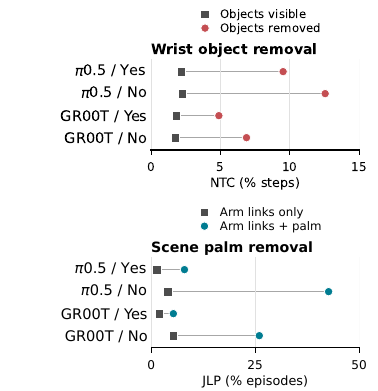}
    \caption{Physical outcomes.}
    \label{fig:content-physical-outcomes}
    \label{fig:content-removal-ntc}
    \label{fig:content-removal-jlp}
  \end{subfigure}
  \caption{Task and physical outcomes of selective content removal on Spatial (150 episodes per cell). (a,b) Task success after removal at episode start. (c) Top: NTC under wrist-object removal versus rendering control. Bottom: JLP after removing scene arm links alone or together with the palm. Both physical indicators use episode-start interventions and up to 105 executed steps. Yes/No and $+$P/$-$P indicate policies with/without proprioception. Exact success values appear in Table~\ref{tab:content-removal}; measurement protocols appear in Appendix~\ref{sec:physical-measurement-details}.}
  \label{fig:content-removal-success}
  \label{fig:content-removal}
  \label{fig:wrist-content-removal}
\end{figure}

\subsection{Different Views Rely on Different Visual Content}

We examine which robot and object depictions support task completion and how changing them affects physical outcomes. We selectively remove the whole robot, individual robot components (arm links, gripper palm, or fingers), or objects from one camera view while leaving the other view unchanged. We implement this in the simulator by disabling rendering of the selected geometries for that camera only; their physics and collisions are unaffected. These interventions begin at episode start unless otherwise noted and are compared against the unmodified rendering.

\textbf{Scene and wrist views rely on different robot cues.}
In the scene view, the gripper palm provides an important visual cue for both \pifive{} and GR00T. For \pifive{}, retaining the gripper palm as the only visible robot component in the scene view preserves 75.3\% task success. In contrast, the wrist view shows a different pattern; removing the fingers reduces task success to 25.3\% for \pifive{} and 12.7\% for GR00T, while scene-view finger removal leaves success largely unchanged.

\textbf{Wrist-view object information must remain available.}
Removing objects from the scene view reduces task success from 94.7--95.3\% to 36.0--57.3\% across the two models, whereas removing them from the wrist view reduces success to 0--2.7\% (Table~\ref{tab:content-removal}). Although the remaining scene view still shows the objects, the policies do not effectively compensate for the loss of wrist-view object information. Wrist-view object removal also increases \texttt{NTC} in both models (Figure~\ref{fig:content-removal-ntc}, top), showing that its consequences extend beyond task failure.

\textbf{Proprioception improves task success and physical outcomes in different ways.}
Proprioception improves both task success and joint behavior when robot depictions are removed. For \pifive{}, removing the scene-view arm links and gripper palm yields 18.7\% success without proprioception and 63.3\% with it (Table~\ref{tab:content-removal}), while \texttt{JLP} decreases from 42.7\% to 8.0\% of episodes (Figure~\ref{fig:content-removal-jlp}, bottom). Under wrist-view object removal, however, success remains at 2.7\% for \pifive{} and 0\% for GR00T, with or without proprioception. Despite unchanged success, proprioception reduces \texttt{NTC} from 12.6\% to 9.5\% of steps for \pifive{}, with a similar reduction for GR00T (Figure~\ref{fig:content-removal-ntc}, top). Thus, proprioception can reduce adverse physical interactions even when it does not improve task success.

\section{Improving Fault Responses and Measuring What Remains}

Our analyses raise the question of whether these vulnerabilities to visual faults can be mitigated. We examine two approaches, one training-based and the other training-free.

\begin{figure}[t]
  \centering
  \begin{subfigure}[t]{.48\linewidth}
    \centering
    \includegraphics[width=\linewidth]{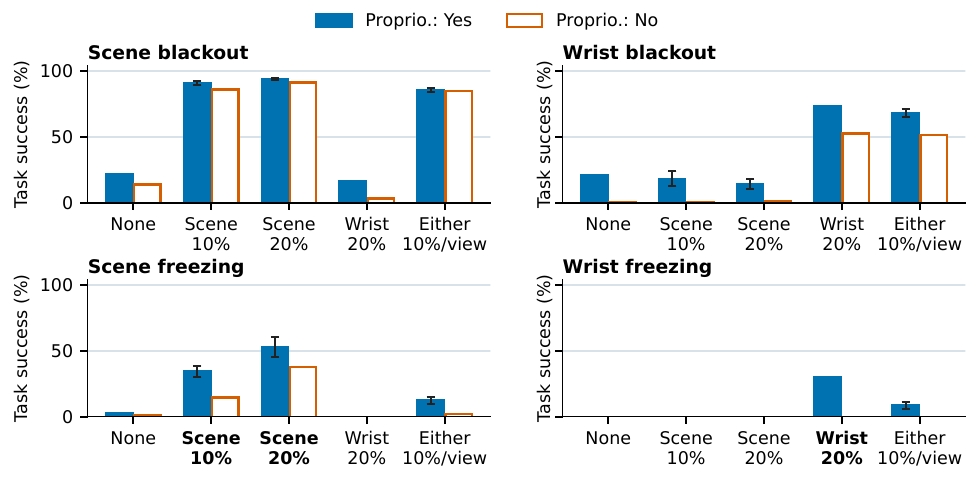}
    \caption{Task success and transfer to freezing.}
    \label{fig:blackout-transfer}
  \end{subfigure}\hfill
  \begin{subfigure}[t]{.48\linewidth}
    \centering
    \includegraphics[width=\linewidth]{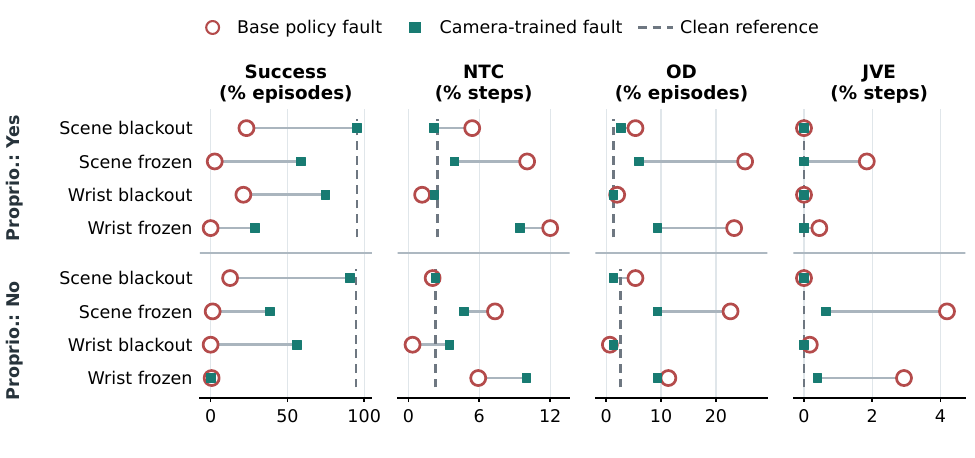}
    \caption{Task success and physical indicators.}
    \label{fig:physical-recovery}
  \end{subfigure}
  \caption{Effects of blackout training on \pifive{} on Spatial, with 150 episodes per condition per run. (a) Whiskers show observed ranges across two training runs where available. Full results appear in Table~\ref{tab:mitigation-training-matrix}. (b)  Physical indicators use the first 105 executed steps; \texttt{NTC} and \texttt{JVE} are step percentages, and \texttt{OD} is an episode percentage.}
  \label{fig:blackout-training-outcomes}
\end{figure}

\subsection{Training-Based Mitigation: Camera Blackout Training}
One straightforward approach is to introduce visual faults during training through image augmentation. We train \pifive{} with camera blackout augmentation, replacing a camera image with a black frame in 20\% of training samples. We evaluate across camera views and proprioception conditions and find that blackout training improves task success in most settings (Figure~\ref{fig:blackout-transfer}). Only under wrist freezing, performance remains low even after blackout training.

\textbf{Blackout training partially transfers to freezing.}
Camera-specific blackout training improves task success from 2.7\% to 53.0\% under scene freezing and from 0.0\% to 30.0\% under wrist freezing, indicating that its benefits extend to frozen inputs.

\textbf{Matching attention allocation yields limited recovery.}
Under blackout of the camera used for augmentation, trained policies allocate more action-expert attention to the unaffected view (Appendix Figure~\ref{fig:app-attention-training-input-mass}). Motivated by this shift, we reweight attention to match selected relative allocations among state and camera tokens measured in trained policies. Despite matching these ratios on fixed probes, scene-blackout success increases only from 22.0\% to 26.7\%, compared with 94.7\% for the trained reference (Appendix~\ref{sec:additional-attention-reweighting}). Whether more targeted interventions can turn this attention shift into reliable recovery remains an open question.

\textbf{Blackout training reduces some physical failure but worsens others.}
With proprioception, scene-blackout training reduces \texttt{OD} under scene freezing from 25.3\% to 6.0\% of episodes, though not to the clean level (1.3\%). Without proprioception, wrist-blackout training reduces \texttt{JVE} under wrist freezing from 2.93\% to 0.39\% of steps but increases \texttt{NTC} from 5.91\% to 9.98\% (Figure~\ref{fig:physical-recovery}). Mitigation should therefore be evaluated across multiple physical indicators, since an improvement in one does not ensure improvement in others.

Overall, blackout training substantially restores task success under blackout and partially under freezing. However, it neither eliminates physical failures nor reduces them uniformly, and wrist freezing remains difficult, especially without proprioception. These limitations motivate future mitigation methods that address faulty visual inputs while jointly targeting task recovery and reductions in adverse physical outcomes. Because blackout training requires retraining the policy, we next test a training-free alternative that replaces the faulty camera input at inference time.

\begin{figure}[t]
  \centering
  \begin{subfigure}[t]{.48\linewidth}
    \centering
    \includegraphics[width=\linewidth]{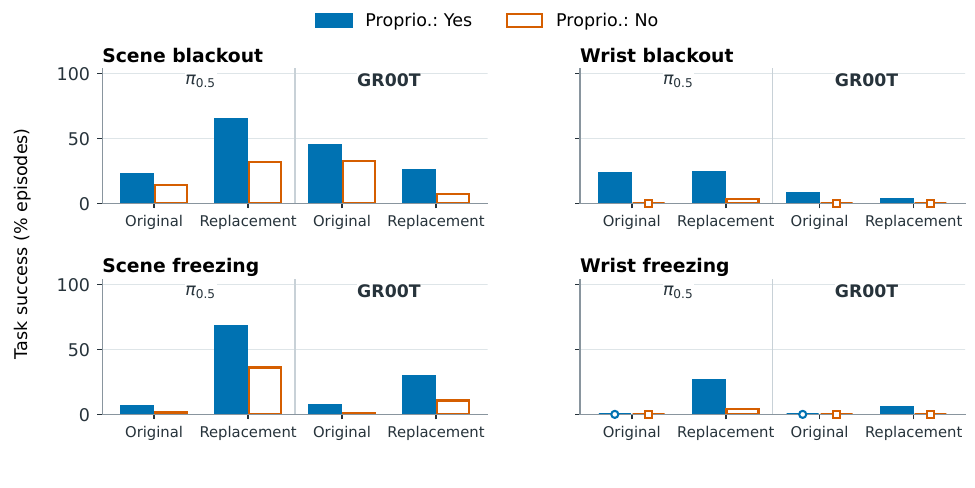}
    \caption{Task success.}
    \label{fig:trainfree-success}
  \end{subfigure}\hfill
  \begin{subfigure}[t]{.48\linewidth}
    \centering
    \includegraphics[width=\linewidth]{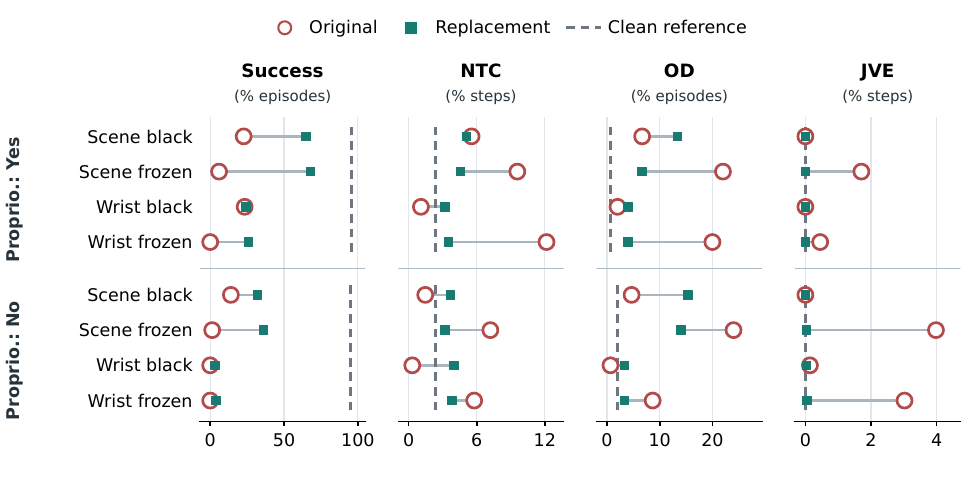}
    \caption{Physical indicators.}
    \label{fig:trainfree-physical}
  \end{subfigure}
  \caption{Mean-embedding replacement under episode-start camera faults on Spatial (150 episodes per arm). (a) Task success for \pifive{} and GR00T. (b) Physical outcomes for \pifive{}.}
  \label{fig:trainfree-embedding-onsets}
\end{figure}
\subsection{Train-Free Mitigation: Mean Embedding Replacement}
\label{sec:trainfree}

Blackout and freezing motivate replacement for different reasons. Under freezing, the camera continues to supply a particular past scene that may no longer match the current state. Replacing its embeddings removes that scene-specific content without attempting to reconstruct the missing current view. Under blackout, current scene information is already absent; replacement instead changes how this absence is represented, substituting a reference derived from normal training images for the representation produced by an all-black image.

For both faults, we use camera-specific, per-token means computed over uncorrupted training images. Among constant replacements, these means minimize the average squared distance to the training embeddings, providing a data-derived reference while preserving the camera's token layout. We evaluate whether this substitution improves task success and physical outcomes when the policy must act using the remaining inputs.

\textbf{Task recovery depends on the camera and policy.}
As shown in Figure~\ref{fig:trainfree-success}, under scene-camera faults, mean-embedding replacement improves task success for \pifive{}. For GR00T, however, its effect depends on the fault type: success increases under freezing and decreases under blackout. Recovery under wrist-camera faults is more limited. Among wrist-camera faults, replacement is significantly effective under freezing for \pifive{}, but the effect is limited in other cases.

\textbf{Physical improvements depend on the fault type.} 
For \pifive{}, replacement reduces \texttt{NTC}, \texttt{OD}, and \texttt{JVE} under episode-start freezing in either camera, with and without proprioception (Figure~\ref{fig:trainfree-physical}). For example, without proprioception, \texttt{JVE} under scene freezing falls from 3.98\% to 0.03\% of steps. Under blackout, however, \texttt{OD} increases in both cameras and proprioception conditions. Even when replacement raises scene-blackout success from 22.7\% to 64.7\% with proprioception, \texttt{OD} rises from 6.7\% to 13.3\% of episodes. One possible explanation is that blackout suppresses movement (Table~\ref{tab:fault-onset-joint-limit}), limiting undesirable contact and object disturbance, while the replacement embedding may be treated by the policy as a valid observation and used for action selection despite the lack of current visual information from that camera. This could increase \texttt{OD} and, in some conditions, \texttt{NTC}. 

Overall, mean-embedding replacement shows that changing how a faulty camera is represented can improve execution without restoring its current visual information. The inconsistent gains of a fixed mean suggest investigating how replacement representations should be selected. Future work could test whether conditioning the replacement on the remaining camera view and available proprioception leads to more reliable recovery.

\section{Related Work}
\textbf{Visual robustness and safety.}
LIBERO-Plus evaluates visual perturbations and uses black-frame ablations to show that wrist observations can partially support task completion when the scene view is unavailable~\citep{Fei_2026_CVPR}. SafeVLA-Bench and ForesightSafety-VLA assess physical risks beyond task success, including contact, object handling, and responses to visual degradation~\citep{fan2026safevlabenchbenchmarksuccesssafetygap,lyu2026foresightsafetyvlaunifieddiagnosticsafety}. Building on these findings, we contrast missing and outdated information through single-camera blackout and freezing at different fault onsets. Our comparisons characterize how these faults produce different executed commands and physical outcomes, even when task-success rates are similarly low.

\textbf{Visual content and proprioception.}
VLA-Trace masks targets, grippers, robots, and backgrounds to identify visual dependencies~\citep{shi2026vlatracediagnosingvisionlanguageactionmodels}. GAP addresses proprioceptive dominance during motion-transition phases~\citep{lu2026when}, while ReViP mitigates false completion through vision--proprioception rebalancing~\citep{li2026revipmitigatingfalsecompletion}. We examine camera-specific dependence on robot and object depictions and whether proprioception compensates for their removal. Beyond task success, we measure how these interventions affect physical indicators.

\textbf{Mitigating observation failures.}
RoboPanoptes uses camera-dropout training for sensor resilience~\citep{XuX-RSS-25}, and RobustVLA trains against multimodal perturbations~\citep{guo2026on}. Without retraining, ActFovea checks visual--state--action consistency to mitigate visual delay and invoke safe failure under frozen replay~\citep{yu2026actfovearuntimesafeguardingvla}. We evaluate camera-specific blackout training and inference-time mean-embedding replacement, focusing on transfer from blackout to freezing, camera-dependent recovery, and whether task-success gains also improve physical outcomes. This approach reveals both the benefits and limitations of adaptation and input replacement under persistent single-camera faults.

\section{Conclusion}
We studied how camera faults affect task performance and physical execution in \pifive{} and GR00T. Blackout and freezing produce distinct physical failure modes that depend on the affected camera and fault timing. Selective interventions reveal strong dependence on wrist-view object information and show that proprioception partly compensates for removed robot depictions but does not restore task success when wrist-view object information is removed, even with the scene view available. Across proprioception and mitigation experiments, improvements in task success do not consistently coincide with reductions in physical risk. These findings suggest that neither the availability of other sensory inputs nor recovery of task performance is sufficient to establish a safe fault response. Safer policies should adapt their behavior to the robot and object information that remains available, accounting for the physical risks of continuing or changing the ongoing execution.

\section{AI disclosure}
In this work, we used generative AI tools to develop theoretical models or conceptual frameworks, propose or refine hypotheses, design or provide feedback on research methodology or experiments, implement methods, assist with translation, support qualitative and thematic data analysis, finding related works and interpret results.

We have not used generative AI tools to generate synthetic datasets, formulate mathematical claims, provide critical ingredients for proving mathematical claims, assist in the writing of proofs, or clean and reformat datasets.

AI-assisted outputs were manually reviewed by one author. The author reviewed research ideas, experimental designs, code, analyses, interpretations, and manuscript text for correctness, and revised when necessary. We take responsibility for the final content of this work, including text, claims or artifacts produced with the aid of generative AI.

\section{Reproducibility Statement}
We describe the camera-fault interventions and evaluation setup in the main text. The appendix provides details on policy interfaces, evaluation cohorts, physical metric definitions, measurement-window sensitivity analyses, and additional experimental results. Code for reproducing our experiments is available at \url{https://anonymous.4open.science/r/blackout-vs-freeze-vla}.

\FloatBarrier
\bibliography{references}
\bibliographystyle{iclr2027_conference}
\clearpage

\appendix
\setlength{\floatsep}{10pt plus 2pt minus 2pt}
\setlength{\textfloatsep}{10pt plus 2pt minus 2pt}
\setlength{\intextsep}{10pt plus 2pt minus 2pt}
\renewcommand{\topfraction}{0.95}
\renewcommand{\bottomfraction}{0.95}
\renewcommand{\textfraction}{0.05}
\renewcommand{\floatpagefraction}{0.75}
\setcounter{topnumber}{4}
\setcounter{bottomnumber}{4}
\setcounter{totalnumber}{6}
\makeatletter
\setlength{\@fptop}{0pt}
\setlength{\@fpsep}{12pt}
\setlength{\@fpbot}{0pt plus 1fil}
\makeatother

\section{Experimental and Training Settings}
\label{sec:experimental-training-settings}
\label{sec:state-interfaces}

\FloatBarrier
\subsection{Training Data and Policy Checkpoints}

Our simulation training set combines LIBERO-Spatial, Object, Goal, and Long: 1,693 episodes with 273,465 frames. For \pifive{}, we use \texttt{physical-intelligence/libero}; for GR00T, we use the same four suites in the model's native training format. Training settings are listed in Table~\ref{tab:appendix-training-settings}.

\begin{table}[!htbp]
\caption{Training settings for the principal simulation policies. The with- and without-proprioception policies are trained separately within each model family. These settings do not describe the released or short-budget checkpoints used in the additional mitigation experiments.}
\label{tab:appendix-training-settings}
\centering\footnotesize
\setlength{\tabcolsep}{4pt}
\renewcommand{\arraystretch}{1.12}
\begin{tabularx}{\linewidth}{@{}l>{\raggedright\arraybackslash}X>{\raggedright\arraybackslash}X@{}}
\toprule
Setting & \pifive{} & GR00T \\
\midrule
Initialization & Base \pifive{} weights & GR00T-N1.7-3B base weights \\
Evaluated update & 10,000 & 4,000 \\
Effective batch size & 32 & 640 \\
Optimizer & AdamW & AdamW \\
Learning rate & Linear warmup to $5\times10^{-5}$ & $10^{-4}$ peak, cosine schedule \\
Warmup & 10,000 updates & 200 updates (5\% of training) \\
Adam $(\beta_1,\beta_2)$ & $(0.9,0.95)$ & $(0.9,0.999)$ \\
Adam $\epsilon$ & $10^{-8}$ & $10^{-8}$ \\
Weight decay & $10^{-10}$ & $10^{-5}$ \\
Gradient-norm clipping & 1.0 & 1.0 \\
Weight averaging & EMA, decay 0.999 & No EMA checkpoint averaging \\
Training precision & bfloat16 & bfloat16 \\
Training seed & 43 & 42 \\
Predicted action horizon & 10 & 40 \\
State-enabled interface & Discretized state in the prompt & Encoded state features \\
State-disabled training & Omit discretized state input & State-feature dropout probability 1 \\
\bottomrule
\end{tabularx}
\end{table}

\paragraph{Proprioception and parameter updates.}
The state vector contains end-effector position (three coordinates), axis-angle orientation (three), and two finger joint positions. For \pifive{}, we discretize this eight-dimensional vector and include it in the language-model input when proprioception is enabled. The policy without proprioception omits state during both training and evaluation. We fine-tune all parameters in both policies from the same base weights, using the same optimizer, batch size, schedule, and seed. Evaluation uses the EMA checkpoints at 10,000 updates.

GR00T disables proprioception by dropping state features with probability one during training and zeroing them at evaluation. With proprioception, the dropout probability is zero. We freeze the vision encoder and language backbone and fine-tune the action-generation modules, projector, and configured normalization and linear layers.

For GR00T, color jitter uses brightness 0.3, contrast 0.4, saturation 0.5, and hue 0.08. The image processor specifies a $256\times256$ target, a $230\times230$ crop, and a crop fraction of 0.95. Accumulating microbatches of 80 over four steps across two replicas gives an effective batch size of 640.

The policies with and without proprioception are trained separately, with one seed per condition. Episode-level intervals describe uncertainty for these checkpoints; they do not capture variation across training runs.

\FloatBarrier
\subsection{Closed-Loop Simulation Evaluation}

Each policy receives a task instruction, a third-person scene image, a wrist image, and state when enabled. It outputs three translation coordinates, three rotation coordinates, and a gripper command. Positive gripper commands close the fingers in the simulator; negative commands open them. We use each model's native action preprocessing and normalization, so command magnitudes are not directly comparable across models.

\begin{table}[!htbp]
\caption{Closed-loop simulation evaluation settings. Episode limits apply before any separately requested post-completion recording. Physical measurements use the windows specified in Appendix~\ref{sec:physical-measurement-details}.}
\label{tab:appendix-evaluation-settings}
\centering\small
\setlength{\tabcolsep}{5pt}
\begin{tabularx}{\linewidth}{@{}l>{\raggedright\arraybackslash}X@{}}
\toprule
Setting & Value \\
\midrule
Control frequency & 20\,Hz \\
Initial settling & Ten simulator steps before policy execution \\
Replanning interval & Five executed actions for \pifive{}; eight for GR00T \\
Denoising steps & Ten for \pifive{}; four for GR00T \\
Episode limit & Spatial: 220; Object: 280; Goal: 300 executed steps \\
Default evaluation size & Ten tasks $\times$ 15 initial states = 150 episodes per suite and condition \\
Pooled success evaluation & Spatial, Object, and Goal; 450 episodes per condition \\
Physical/content evaluation & Spatial, with condition-specific clean or rendering controls \\
Evaluation seed & 7 \\
Success criterion & The simulator's task-completion predicate \\
\bottomrule
\end{tabularx}
\end{table}

Figure~\ref{fig:trained-success} combines success rates from 150 episodes each of Spatial, Object, and Goal, with exact values in Table~\ref{tab:trained-success}. Physical-outcome and content-removal experiments use Spatial and their own clean or rendering controls. Rollouts and fixed-observation probes use separate cohorts, with comparisons made within each cohort.

\begin{table}[!htbp]
\caption{Task success (\%) for policies with and without proprioception. Both models average Spatial, Object, and Goal ($n=450$ per condition).}
\label{tab:trained-success}
\centering
\footnotesize
\setlength{\tabcolsep}{3.5pt}
\begin{tabular}{llrrrrr}
\toprule
Model & Proprio. & Clean & Scene black & Wrist black & Scene frozen & Wrist frozen \\
\midrule
\pifive{} & Yes & 93.1 & 28.9 & 13.1 & 5.8 & 3.6 \\
\pifive{} & No & 94.4 & 9.1 & 1.8 & 0.2 & 0.0 \\
GR00T & Yes & 98.7 & 51.6 & 12.2 & 9.3 & 0.0 \\
GR00T & No & 97.1 & 34.4 & 0.0 & 5.1 & 0.0 \\
\bottomrule
\end{tabular}
\end{table}

\paragraph{Fault application and timing.}
We alter one camera's input while keeping the other image, instruction, and state interface unchanged. Blackout sets the affected RGB image to zero. Freezing repeatedly supplies the image captured when the fault begins, while the physical simulation continues. Episode-start faults take effect at the first policy observation after settling.

For after-closure faults, the finger width must remain below 0.020\,m for three consecutive steps. We then wait a uniformly sampled 0--10 executed steps before applying the fault. Because the trigger uses finger width, it can activate without an object in the gripper. The episode-start and after-closure experiments use separate evaluation sets and clean references.

\paragraph{Selective content removal.}
For content removal, the renderer omits selected arm links, gripper palm, fingers, or objects from one camera view. Collision geometry, physical dynamics, the instruction, and the other camera view remain unchanged. Control rollouts use the same renderer with the selected content visible.

\FloatBarrier
\subsection{Real-Robot Training and Evaluation}

\paragraph{Task and training.}
A WidowX robot with scene and wrist cameras places a black block in a bowl. Training uses 100 clean demonstrations with 19,346 frames sampled at 8\,Hz, without blackout augmentation. Scene and wrist images have native resolutions of $640\times360$ and $640\times480$. State and action vectors contain six arm-joint coordinates and one gripper coordinate; actions are joint-position targets. Both \pifive{} policies start from the same base weights and train for 10,000 updates with seed 43, batch size 32, and EMA decay 0.999. The learning rate warms up over 1,000 updates to $5\times10^{-5}$, then stays constant. AdamW settings follow Table~\ref{tab:appendix-training-settings}; the predicted action horizon is 16. The policy with proprioception receives discretized joint state; the other omits it.

\paragraph{Execution and physical outcomes.}
Each policy runs ten trials under each input condition (clean, scene blackout, scene freezing, wrist blackout, and wrist freezing), for 100 trials in total. Faults last from episode start to termination. At 9\,Hz, the controller executes eight actions per inference and allows up to 1,200 steps (approximately 133\,s). It clips command step sizes and joint positions, with soft arm-joint bounds one degree inside the firmware limits.

The operator stops successful trials after completion and ends some failures before the time limit. A deadman timeout at 130\,s can leave a short inactive interval in trials that reach the step limit.

For each real-robot trial, the operator records whether non-target contact or an object drop occurred. We divide event counts by all ten trials in the condition. These trial-level labels differ from the simulation's contact step rate and automated grasp-loss rule. Joint-limit margin is the smallest angular distance to a soft bound across arm joints and recorded times, excluding the gripper. We report the median margin across trials; simulation JLP instead counts episodes crossing a threshold. Wilson intervals cover binomial trial uncertainty, not variation across training runs.

\FloatBarrier
\section{Physical Measurements and Window Sensitivity}
\label{sec:physical-measurement-details}

We use the physical indicators in Table~\ref{tab:safety-measures}, with the definitions and measurement windows below.

\paragraph{Recording and measurement windows.}
Physical traces are sampled after every executed step at 20\,Hz, missing any force peaks between samples. For the 150 episodes per Spatial condition in Table~\ref{tab:fault-onset-joint-limit}, we measure from the start of the trace through first task completion, or through episode end for failures. This includes pre-fault steps, so observation windows can differ in length. Post-goal continuation is excluded, and task success is recorded for the whole episode.

Drop and the sensitivity analysis include the completion sample. For the other primary indicators, \pifive{} includes that sample but GR00T excludes it.

Selective-removal and mitigation comparisons use up to the first 105 recorded steps, a different exposure window from the camera-fault table. Captions specify step or episode percentages for each indicator.

\paragraph{Contact, object movement, and joint motion.}
\texttt{NTC} counts steps with robot contact involving non-target, non-arena objects. We divide by each episode's measured steps and average the fractions across episodes. Target and table contacts are excluded; fixture and destination-support contacts can count even when intentional. For \texttt{OD}, a movable non-target object must stay more than 2\,cm from its first recorded position for six consecutive samples.

For \texttt{JVE}, any arm joint exceeding its manufacturer's absolute velocity limit qualifies. Table~\ref{tab:fault-onset-joint-limit} counts episodes with at least one exceedance. The blackout-training figures instead average each episode's fraction of affected steps. For \texttt{JLP}, an episode qualifies if J2 or J7 enters a 0.05 margin from either limit in normalized joint position.

At each step, the recorded force is the maximum contact-force norm across robot self-contact and external contact, including normal and tangential components. The contact pair producing this maximum is not recorded. Sustained force ($F_5$) requires this value to exceed 100\,N for at least five consecutive steps. To compute $\|\Delta xyz\|$, we take the norm of the first three executed action coordinates, average over steps within each episode, then average over episodes. Its units are native action units, not physical travel distance.

\paragraph{Target-object drops.}
\texttt{Drop} requires a grasp, loss of finger contact, and rapid descent. Let $z_t$ be the target height and $z_0$ its first recorded height after settling. We establish a grasp when bilateral finger contact and $z_t-z_0>2$\,cm last for three consecutive steps. After that, a candidate loss time $t$ occurs when neither finger contacts the target and $z_t-z_0>2$\,cm still holds. A drop requires
\[
  z_u-z_{u+2}>5\,\mathrm{cm}
  \quad\text{for some }u\in\{t,t+1,t+2,t+3\},
\]
with no finger recontact from $t$ through $u+2$. The descent is measured over two executed steps (0.1\,s), starting no more than three steps after contact loss. We discard candidates if first task completion follows within 15 steps of contact loss. The search ends at first completion for successful episodes and at the end of the trace for failures, excluding post-goal events. Each episode contributes at most one Drop, and the rate uses all 150 episodes as its denominator. This rule can include intentional off-goal placement or exclude a fall followed promptly by completion.

\paragraph{Opportunity to observe an event.}
Episodes that never establish a grasp cannot register a Drop. Table~\ref{tab:drop-grasp-counts} therefore reports grasp counts and Drop rates both overall and among grasping episodes. A grasp counts anywhere in the measurement window, including after fault onset. The grasping subset differs across conditions, so we use the all-episode rate for the main comparison.

Successful episodes often provide fewer observed steps than failures. Pre-fault steps can also dilute the measured rate of events that occur after the fault. Table~\ref{tab:physical-window-exposure} reports durations and exposure counts. Episodes whose closure trigger never fires still enter the primary denominator but contribute no steps to the post-onset window.

\begin{table*}[p]
\centering\scriptsize
\caption{Drop counts and grasp opportunity under camera faults. Every row contains 150 episodes. The reported Drop rate uses all 150 episodes; the conditional rate uses only episodes with an established grasp. A grasp requires bilateral finger contact and target height more than 2\,cm above its first recorded height for three consecutive steps. Camera faults are introduced after gripper closure, with matched clean references.}
\label{tab:drop-grasp-counts}
\setlength{\tabcolsep}{4pt}
\begin{tabular}{lllrrrr}
\toprule
Model & Proprio. & Condition & Drop ($n$) & Grasp ($n$) & Drop (\%) & Drop / grasp (\%) \\
\midrule
\multicolumn{7}{l}{\textit{Camera faults}} \\
\pifive{} & Yes & Clean & 2 & 146 & 1.3 & 1.4 \\
\pifive{} & Yes & Scene blackout & 3 & 137 & 2.0 & 2.2 \\
\pifive{} & Yes & Scene freezing & 8 & 144 & 5.3 & 5.6 \\
\pifive{} & Yes & Wrist blackout & 6 & 135 & 4.0 & 4.4 \\
\pifive{} & Yes & Wrist freezing & 3 & 138 & 2.0 & 2.2 \\
\pifive{} & No & Clean & 2 & 141 & 1.3 & 1.4 \\
\pifive{} & No & Scene blackout & 14 & 125 & 9.3 & 11.2 \\
\pifive{} & No & Scene freezing & 3 & 144 & 2.0 & 2.1 \\
\pifive{} & No & Wrist blackout & 43 & 130 & 28.7 & 33.1 \\
\pifive{} & No & Wrist freezing & 10 & 134 & 6.7 & 7.5 \\
GR00T & Yes & Clean & 3 & 145 & 2.0 & 2.1 \\
GR00T & Yes & Scene blackout & 4 & 140 & 2.7 & 2.9 \\
GR00T & Yes & Scene freezing & 7 & 148 & 4.7 & 4.7 \\
GR00T & Yes & Wrist blackout & 10 & 142 & 6.7 & 7.0 \\
GR00T & Yes & Wrist freezing & 6 & 144 & 4.0 & 4.2 \\
GR00T & No & Clean & 1 & 143 & 0.7 & 0.7 \\
GR00T & No & Scene blackout & 1 & 136 & 0.7 & 0.7 \\
GR00T & No & Scene freezing & 1 & 148 & 0.7 & 0.7 \\
GR00T & No & Wrist blackout & 64 & 142 & 42.7 & 45.1 \\
GR00T & No & Wrist freezing & 10 & 145 & 6.7 & 6.9 \\
\bottomrule
\end{tabular}
\end{table*}

\begin{table}[t]
\centering\scriptsize
\caption{Measurement opportunity in the representative contrasts. Every row has 150 episodes. Prefix is the median number of recorded steps through first completion (full trace for failures). Full post50 counts episodes with all 50 observed steps after logged fault onset before completion; zero post50 counts episodes with no such exposure. Episodes with shorter exposure remain in the all-episode denominator. Grasp is the number satisfying the established-grasp rule before completion.}
\label{tab:physical-window-exposure}
\setlength{\tabcolsep}{4pt}
\begin{tabular}{lllrrrr}
\toprule
Model & Proprio. & Condition & Grasp & Prefix & Full post50 & Zero post50 \\
\midrule
\pifive{} & Yes & Wrist blackout & 135 & 132.0 & 117 & 0 \\
\pifive{} & Yes & Wrist freezing & 138 & 135.0 & 126 & 0 \\
\pifive{} & No & Wrist blackout & 130 & 220.0 & 145 & 1 \\
\pifive{} & No & Wrist freezing & 134 & 220.0 & 136 & 1 \\
GR00T & Yes & Wrist blackout & 142 & 183.5 & 125 & 0 \\
GR00T & Yes & Wrist freezing & 144 & 220.0 & 141 & 0 \\
GR00T & No & Wrist blackout & 142 & 220.0 & 140 & 0 \\
GR00T & No & Wrist freezing & 145 & 220.0 & 142 & 0 \\
\bottomrule
\end{tabular}
\end{table}

\paragraph{Uncertainty for representative contrasts.}
For each of the four policies, we form 150 episode pairs by task and initial-state index, comparing wrist blackout and freezing introduced after gripper closure. We verify matching task labels, evaluation seeds, and first recorded target positions, then compare Drop and JVE within each evaluation set. Table~\ref{tab:physical-window-sensitivity} reports mean paired differences and percentile 95\% bootstrap intervals. We resample the ten tasks with replacement 10,000 times, keeping all 15 paired episodes for each sampled task (bootstrap seed 20260925). These exploratory intervals have no adjustment for multiple comparisons. They apply to the tested checkpoints and simulator setup, without accounting for variation across training runs.

\paragraph{Measurement-window sensitivity.}
To check sensitivity to observation duration, we repeat the comparisons over the first 105 steps, the first 50 steps after logged fault onset, and the full recorded trace. Both short windows end at first completion if it occurs sooner. We retain all pairs, even when fewer steps are available. Drop uses the original height reference and prior grasp history, but both contact loss and the qualifying descent must fall inside the chosen window. The 15-step completion exclusion is unchanged.

Both short windows retain positive blackout-minus-freezing Drop differences for all four policies. For GR00T $-$P, the difference falls from 36.0\% absolute in the primary window to 16.0\% with the 105-step cap and 17.3\% in the post-onset window. The primary intervals for both +P policies include or touch zero. JVE changes more with the window: both \pifive{} differences are zero over the first 50 steps after fault onset, suggesting that some of the primary difference emerges later in execution. Adding the recorded post-goal continuation leaves these contrasts unchanged. We have not checked this for other conditions or metrics.

\begin{table*}[t]
\centering\scriptsize
\caption{Uncertainty and measurement-window sensitivity for representative physical contrasts. Each contrast uses 150 paired Spatial episodes. A and B name the two conditions in each block. Rates are percentages of episodes, and $\Delta=A-B$ is an absolute difference on that scale. Primary intervals are 95\% task-cluster bootstrap intervals from 10,000 resamples of the ten tasks. The primary window ends at first task completion, or at episode end for failures. Additional columns give differences for the first 105 steps, the first 50 steps from logged fault onset (both capped at first completion), and the full recorded trace. Drop uses the same contact-conditioned rule in every column; both contact loss and the qualifying descent must lie inside the selected window.}
\label{tab:physical-window-sensitivity}
\setlength{\tabcolsep}{3pt}
\begin{tabular}{lllrrrrrr}
\toprule
Model & Proprio. & Metric & A & B & Primary $\Delta$ [95\% CI] & $\Delta_{105}$ & $\Delta_{\mathrm{post50}}$ & $\Delta_{\mathrm{full}}$ \\
\midrule
\multicolumn{9}{l}{\textit{A: wrist blackout; B: wrist freezing (after gripper closure)}} \\
\pifive{} & Yes & \texttt{Drop} & 4.0 & 2.0 & +2.0 [+0.0, +4.7] & +2.7 & +2.7 & +2.0 \\
\pifive{} & Yes & \texttt{JVE} & 0.0 & 6.7 & -6.7 [-14.0, -0.7] & +0.0 & +0.0 & -6.7 \\
\pifive{} & No & \texttt{Drop} & 28.7 & 6.7 & +22.0 [+8.0, +38.7] & +4.7 & +6.7 & +22.0 \\
\pifive{} & No & \texttt{JVE} & 3.3 & 30.7 & -27.3 [-52.7, -4.7] & -0.7 & +0.0 & -27.3 \\
GR00T & Yes & \texttt{Drop} & 6.7 & 4.0 & +2.7 [-2.0, +7.3] & +2.0 & +2.0 & +2.7 \\
GR00T & Yes & \texttt{JVE} & 0.0 & 26.7 & -26.7 [-48.0, -8.0] & -0.7 & -3.3 & -26.7 \\
GR00T & No & \texttt{Drop} & 42.7 & 6.7 & +36.0 [+24.0, +48.0] & +16.0 & +17.3 & +36.0 \\
GR00T & No & \texttt{JVE} & 2.0 & 58.7 & -56.7 [-80.0, -32.7] & -5.3 & -4.7 & -56.7 \\
\bottomrule
\end{tabular}
\end{table*}

\FloatBarrier
\section{Additional Action Analyses}
\label{sec:app-action-context}

\subsection{Full-Action Rollout Context}
\begin{figure}[!htbp]
  \centering
  \begin{subfigure}[t]{.49\linewidth}
    \centering
    \includegraphics[width=\linewidth]{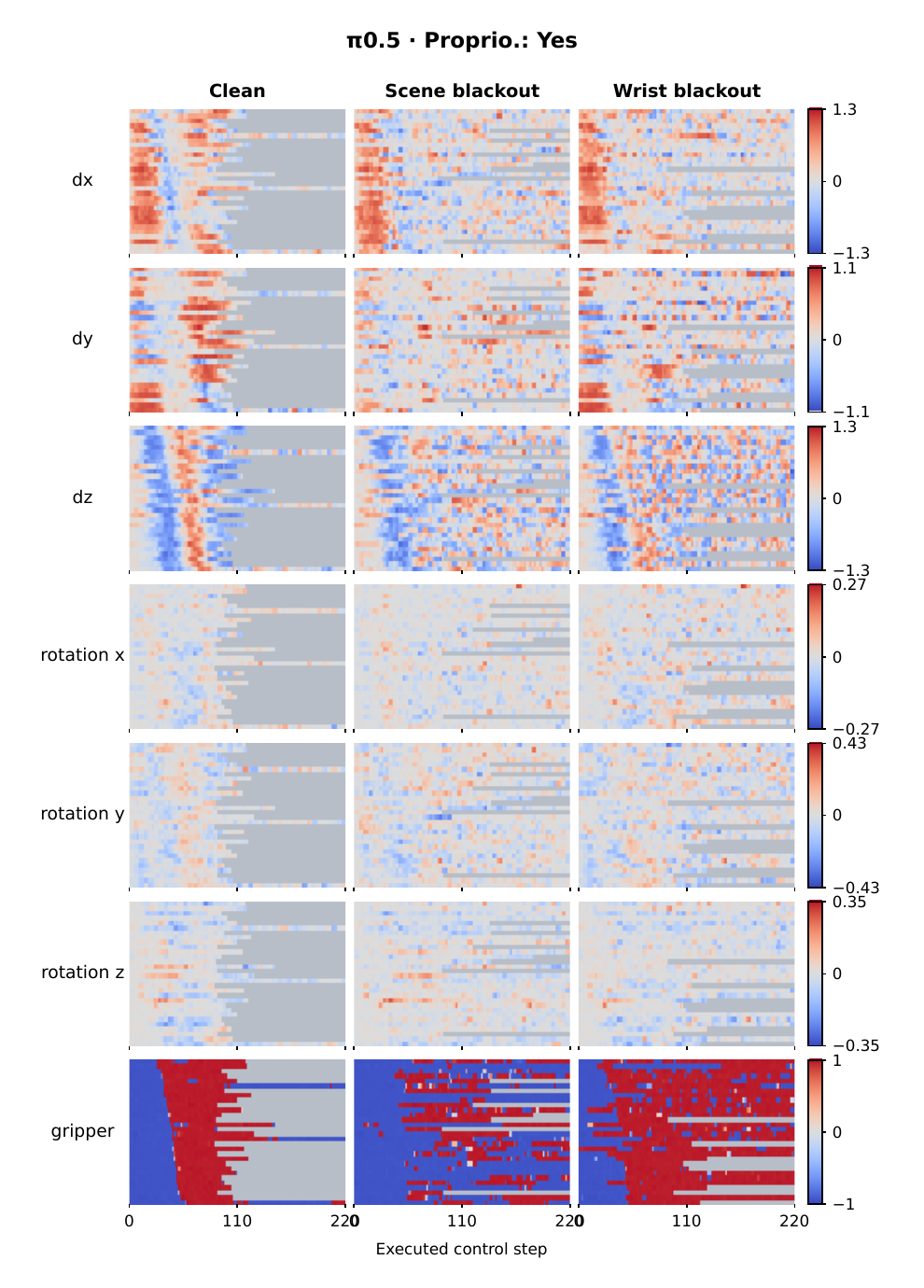}
    \caption{With proprioception.}
    \label{fig:pi-full-action}
  \end{subfigure}\hfill
  \begin{subfigure}[t]{.49\linewidth}
    \centering
    \includegraphics[width=\linewidth]{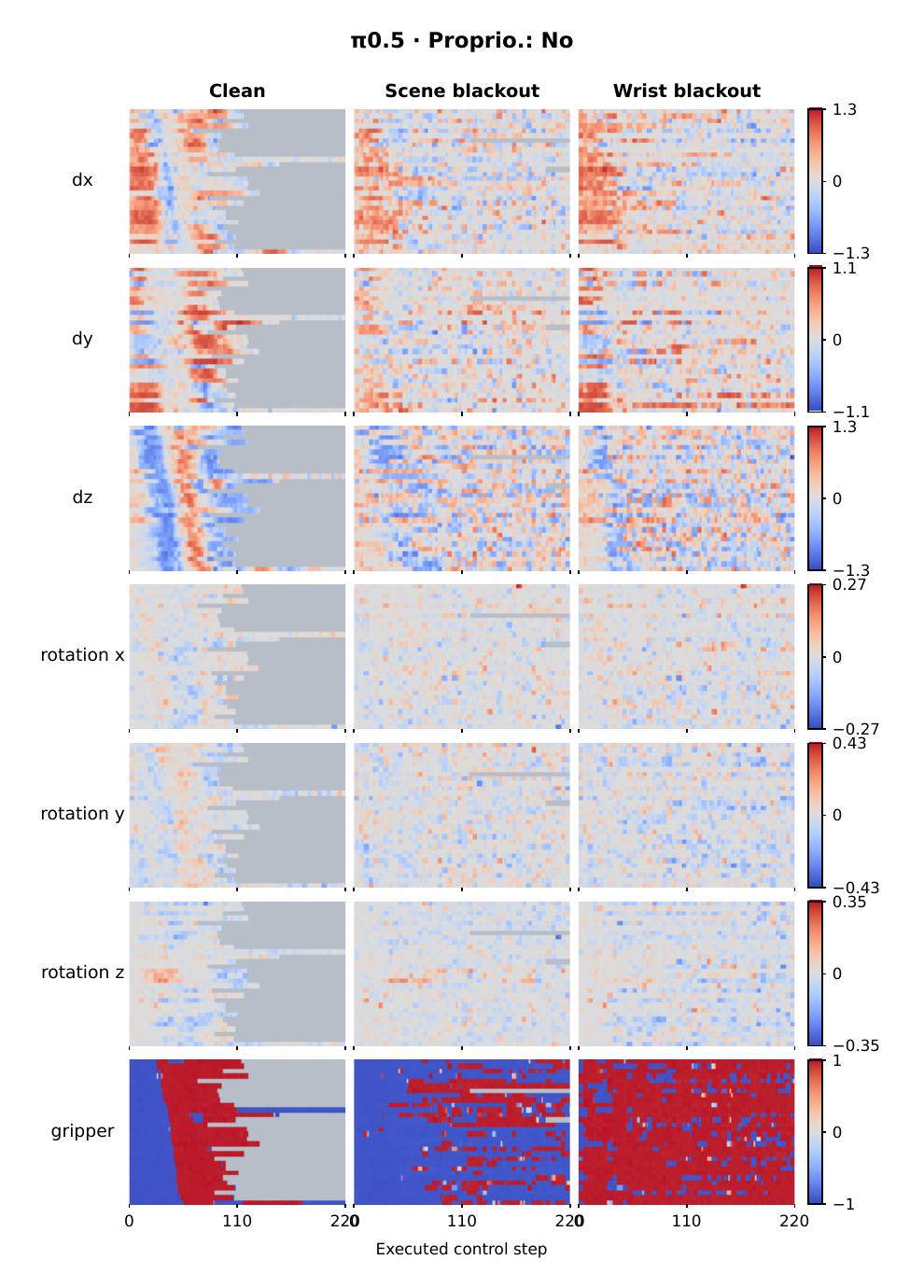}
    \caption{Without proprioception.}
    \label{fig:pi-full-action-no}
  \end{subfigure}
  \caption{\pifive{} executed action coordinates under clean inputs, scene blackout, and wrist blackout. Rows follow the clean first-close order; gray marks time after episode termination. Color scales are shared across conditions and panels for each coordinate.}
  \label{fig:pi-full-actions}
\end{figure}

\begin{figure}[!htbp]
  \centering
  \begin{subfigure}[t]{.49\linewidth}
    \centering
    \includegraphics[width=\linewidth]{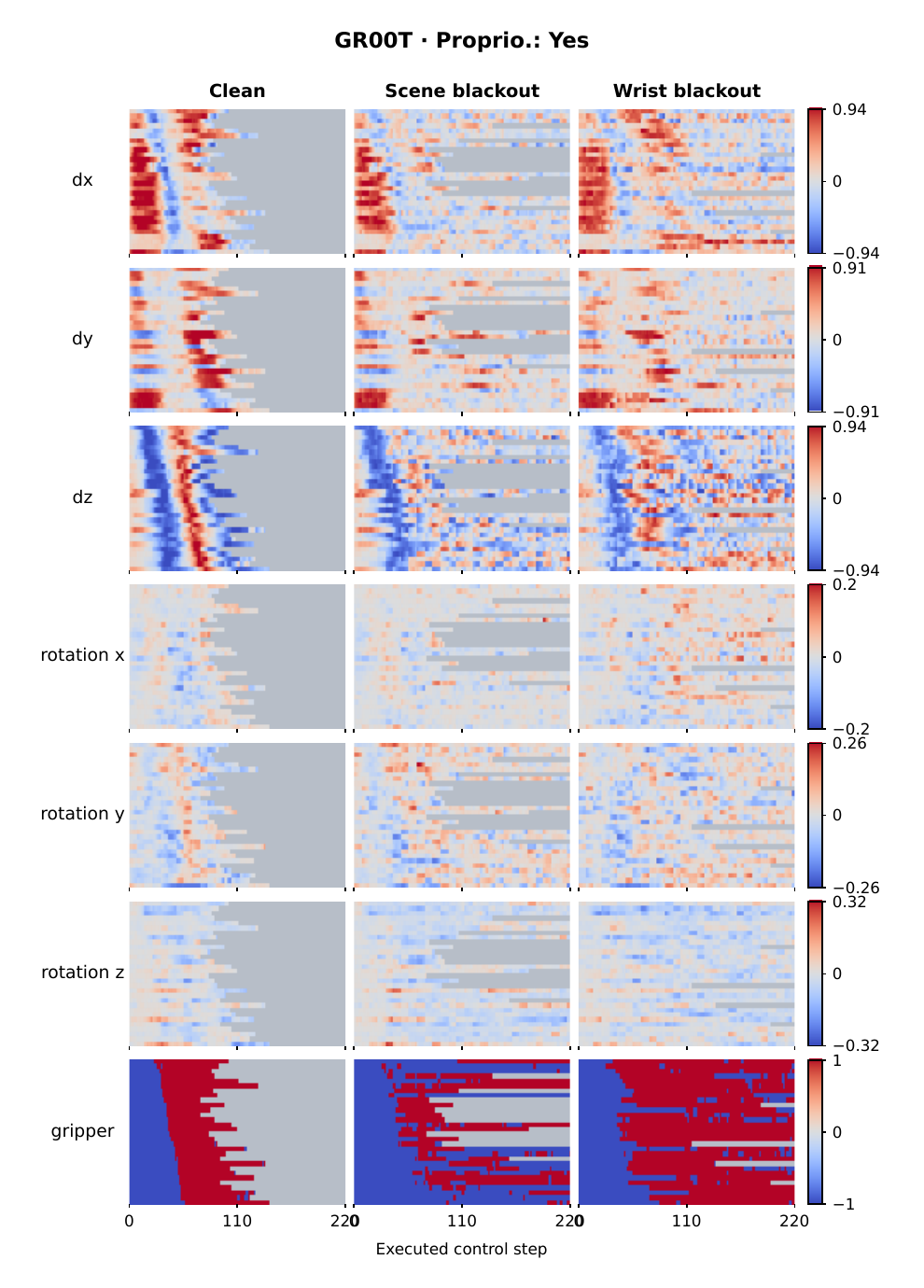}
    \caption{With proprioception.}
    \label{fig:groot-full-action}
  \end{subfigure}\hfill
  \begin{subfigure}[t]{.49\linewidth}
    \centering
    \includegraphics[width=\linewidth]{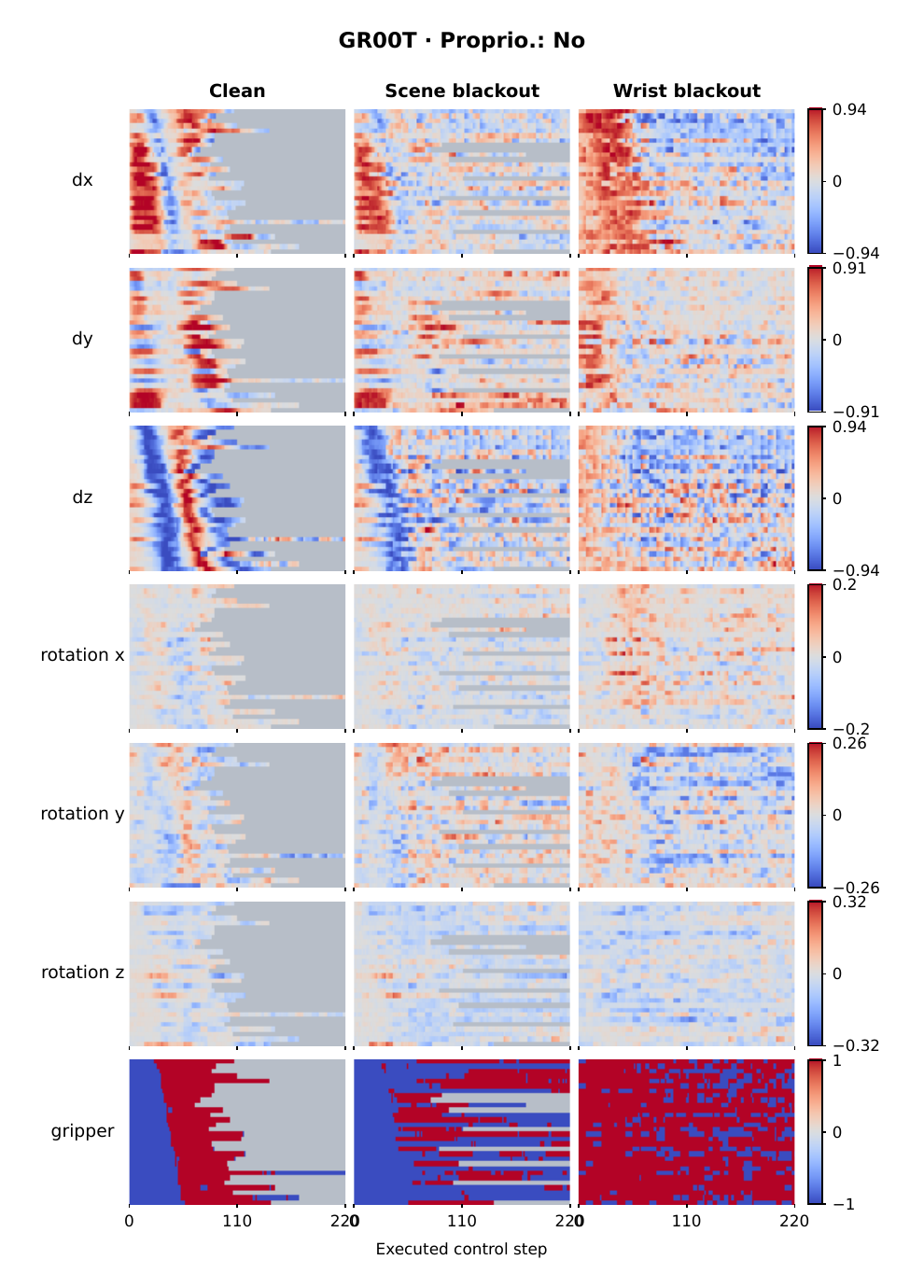}
    \caption{Without proprioception.}
    \label{fig:groot-full-action-no}
  \end{subfigure}
  \caption{GR00T executed action coordinates under clean inputs, scene blackout, and wrist blackout. Rows follow the clean first-close order; gray marks time after episode termination. Color scales are shared across conditions and panels for each coordinate.}
  \label{fig:groot-full-actions}
\end{figure}

Figures~\ref{fig:pi-full-actions} and~\ref{fig:groot-full-actions} extend the gripper raster to all seven executed action coordinates. Each policy uses the same 30 selected episodes across clean and blackout conditions. The plots show native command values, not measured displacement or contact force.

\FloatBarrier
\subsection{Executed-Command Direction}
\label{sec:command-direction}

Table~\ref{tab:command-direction-windows} compares early and later command directions and magnitudes against clean execution, alongside the physical measurements in Appendix~\ref{sec:physical-measurement-details}.

\begin{table}[!htbp]
\caption{Median xyz command cosine relative to paired clean execution. Each cell shows early $[0,20)$ $\rightarrow$ later $[50,80)$ values over eligible episode-step pairs at $\epsilon=0.05$. A value near 1 indicates similar native command direction, irrespective of magnitude.}
\label{tab:command-direction-windows}
\centering\small
\begin{tabular}{llrrrr}
\toprule
Policy & Proprio. & Scene black & Scene frozen & Wrist black & Wrist frozen \\
\midrule
\pifive{} & Yes & $.94\to.27$ & $1.00\to.33$ & $.99\to.74$ & $1.00\to-.10$ \\
\pifive{} & No & $.85\to.15$ & $1.00\to.24$ & $.97\to.55$ & $.99\to.11$ \\
GR00T & Yes & $.99\to.46$ & $1.00\to.33$ & $.98\to.76$ & $.99\to.07$ \\
GR00T & No & $.99\to.31$ & $1.00\to.12$ & $.50\to.06$ & $.99\to.13$ \\
\bottomrule
\end{tabular}
\end{table}

\subsection{Immediate Gripper Responses to Visual Removal}
\label{sec:immediate-gripper-response}

\begin{figure}[!htbp]
  \centering
  \includegraphics[width=\linewidth]{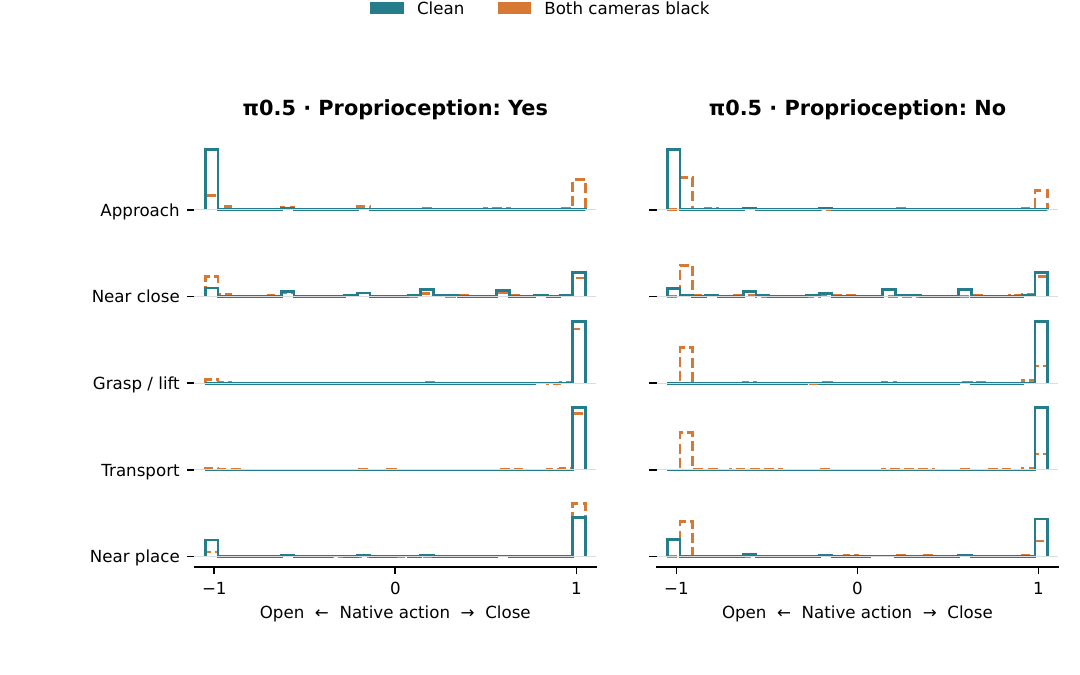}
  \caption{Immediate \pifive{} gripper commands for 200 paired Spatial observations per phase, comparing clean inputs with both camera images blacked out. Commands are averaged over the planned five-step execution prefix. Open is on the left and close is on the right.}
  \label{fig:phase-gripper-distributions}
\end{figure}

Figure~\ref{fig:phase-gripper-distributions} measures \pifive{}'s immediate gripper response to blacking out both cameras at fixed observations. Without proprioception, command distributions shift substantially from clean inputs at the same task phase; with proprioception, they stay closer. The main-text rollouts instead track execution under single-camera faults.

\FloatBarrier
\section{Selective Visual-Content Interventions}
\label{sec:selective-content-details}

\subsection{Exact Success Rates}
Table~\ref{tab:content-removal} gives the detailed removal results corresponding to Figure~\ref{fig:content-removal}. Interventions begin at episode start and use the corresponding rendering controls.

\begin{table}[!htbp]
\caption{Exact task success for the selective-removal conditions in Figure~\ref{fig:content-removal}. Cells show success percentage (successful episodes), with 150 LIBERO-Spatial episodes per condition. Robot rendering controls remove shadows; object rendering controls retain them. The other camera is unmodified.}
\label{tab:content-removal}
\centering\small
\setlength{\tabcolsep}{3pt}
\begin{tabular}{@{}lrrrr@{}}
\toprule
 & \multicolumn{2}{c}{\pifive{}} & \multicolumn{2}{c}{GR00T} \\
\cmidrule(lr){2-3}\cmidrule(l){4-5}
Proprioception & Yes & No & Yes & No \\
\midrule
\multicolumn{5}{@{}l}{\textit{Scene view}} \\
Robot rendering control & 95.3 (143) & 94.0 (141) & 97.3 (146) & 94.0 (141) \\
Fingers hidden & 95.3 (143) & 96.0 (144) & 96.0 (144) & 96.7 (145) \\
Arm links hidden & 93.3 (140) & 90.7 (136) & 96.7 (145) & 88.7 (133) \\
Arm links and fingers hidden & 89.3 (134) & 75.3 (113) & 92.7 (139) & 81.3 (122) \\
Arm links and palm hidden & 63.3 (95) & 18.7 (28) & 74.0 (111) & 44.0 (66) \\
Whole robot hidden & 50.7 (76) & 13.3 (20) & 46.7 (70) & 23.3 (35) \\
\addlinespace[2pt]
Object rendering control & 94.7 (142) & 95.3 (143) & 96.7 (145) & 94.7 (142) \\
Objects hidden & 42.0 (63) & 36.0 (54) & 82.7 (124) & 57.3 (86) \\
\midrule
\multicolumn{5}{@{}l}{\textit{Wrist view}} \\
Robot rendering control & 96.0 (144) & 97.3 (146) & 96.7 (145) & 99.3 (149) \\
Fingers hidden & 59.3 (89) & 25.3 (38) & 88.0 (132) & 12.7 (19) \\
Whole robot hidden & 49.3 (74) & 6.7 (10) & 73.3 (110) & 4.7 (7) \\
\addlinespace[2pt]
Object rendering control & 95.3 (143) & 93.3 (140) & 98.7 (148) & 97.3 (146) \\
Objects hidden & 2.7 (4) & 2.7 (4) & 0.0 (0) & 0.0 (0) \\
\bottomrule
\end{tabular}
\end{table}

\FloatBarrier
\subsection{Physical Outcomes and Executed Commands}
Table~\ref{tab:content-joint-outcomes} reports task success, translation-command changes, and physical outcomes for selective content removal.

\begin{table}[!htbp]
\centering
\caption{Task, command, and physical outcomes after selective content removal. Translation reports RMS change from the matched rendering control over the first 20 executed steps. Physical indicators use up to 105 steps: NTC is non-target contact, and OD is sustained non-target object displacement. Each row contains 150 episodes.}
\label{tab:content-joint-outcomes}
\small
\setlength{\tabcolsep}{3.6pt}
\begin{tabular}{llcrrrr}
\toprule
Removed content & Policy & Proprio. & Success & $\Delta xyz$ & NTC & OD \\
\midrule
Scene robot & \pifive{} & Yes  & .507 & .034 & .055 & .133 \\
            & \pifive{} & No & .133 & .084 & .082 & .247 \\
            & GR00T     & Yes  & .467 & .049 & .029 & .073 \\
            & GR00T     & No & .233 & .075 & .045 & .200 \\
\midrule
Wrist objects & \pifive{} & Yes  & .027 & .061 & .095 & .200 \\
              & \pifive{} & No & .027 & .110 & .126 & .327 \\
              & GR00T     & Yes  & .000 & .144 & .049 & .080 \\
              & GR00T     & No & .000 & .132 & .069 & .140 \\
\bottomrule
\end{tabular}
\end{table}

\FloatBarrier
\section{Camera Disagreement: Protocol and Detailed Results}
\label{sec:camera-disagreement-details}

\subsection{Paired Inputs and Reference Actions}
We compare actions from four scene/wrist image pairs: A/A, B/A, A/B, and B/B. Matching pairs provide reference actions; mixed pairs change one camera. Instruction and any proprioceptive input stay fixed. Actions are queried without execution, measuring which view the policy follows under these image changes, rather than fixed camera weights.

\begin{figure}[!htbp]
  \centering
  \begin{subfigure}[t]{.80\linewidth}
    \centering
    \includegraphics[width=\linewidth]{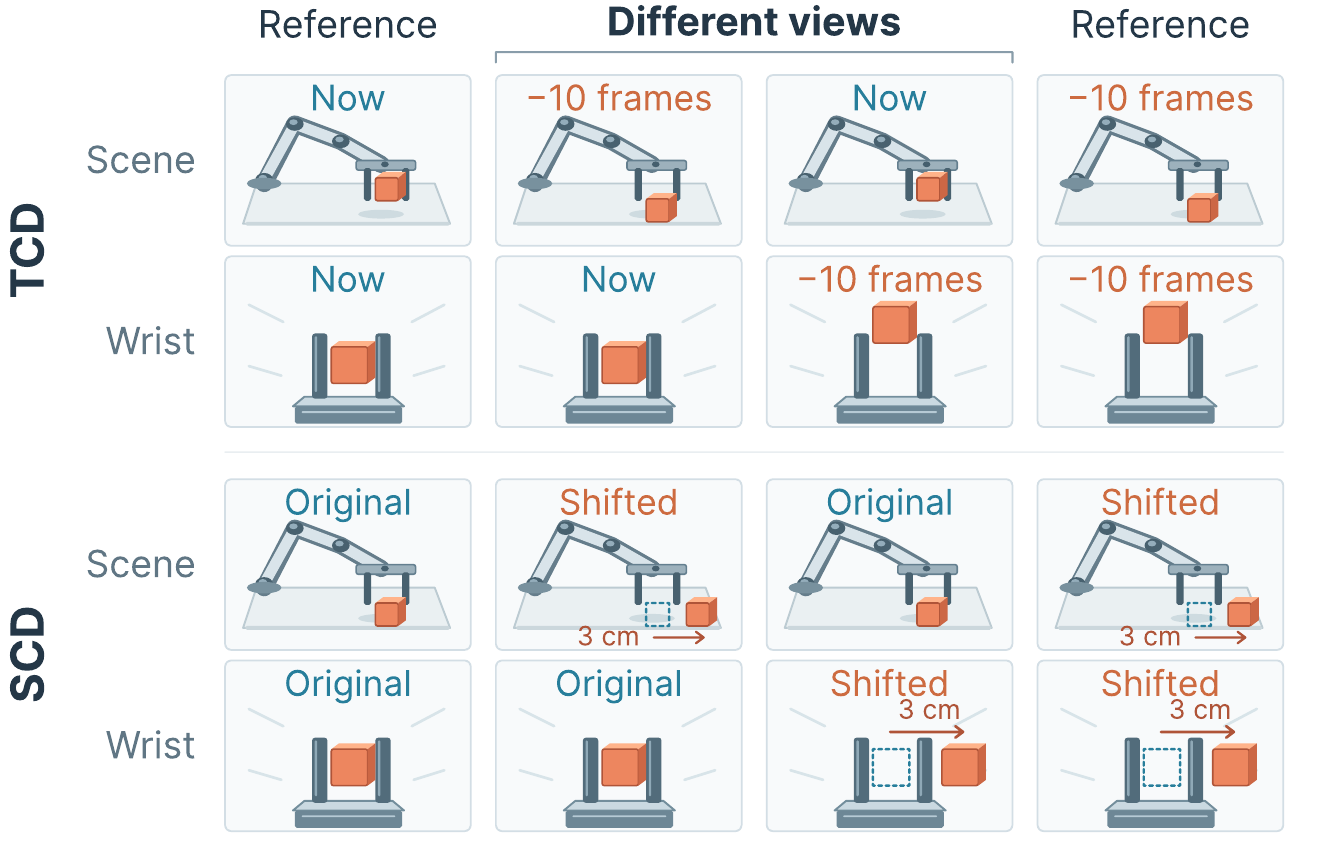}
    \caption{Disagreement setups.}
    \label{fig:camera-conflict-setup}
  \end{subfigure}

  \medskip
  \begin{subfigure}[t]{.47\linewidth}
    \centering
    \includegraphics[width=\linewidth]{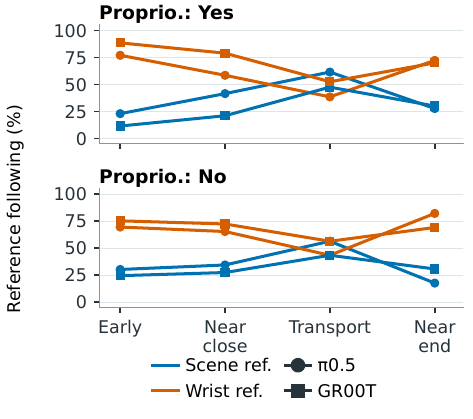}
    \caption{Temporal (10-frame lag).}
    \label{fig:camera-conflict-temporal}
  \end{subfigure}\hfill
  \begin{subfigure}[t]{.47\linewidth}
    \centering
    \includegraphics[width=\linewidth]{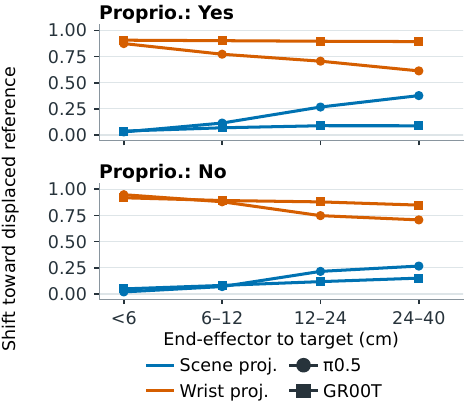}
    \caption{Spatial (3\,cm target shift).}
    \label{fig:camera-conflict-spatial}
  \end{subfigure}
  \caption{Camera-disagreement inputs and policy responses. (a) Scene/wrist input pairs with matching views (outer columns) or different times or target object positions (middle columns). (b) Scene- and wrist-reference following across execution stages, averaged over both disagreement conditions. (c) Median translation-action projections toward the shifted-target reference, grouped by end-effector distance.}
  \label{fig:camera-conflict-results}
\end{figure}

For temporal disagreement, A is the current observation and B is from 10 frames earlier. For spatial disagreement, A shows the original target position and B shifts the target by 3\,cm. Figure~\ref{fig:camera-conflict-results} shows the setups and results. We measure temporal responses as reference-following percentages and spatial responses as continuous translation-action projections.

\subsection{Temporal Reference Following}
Let $\mathbf a_{\mathrm{now}}$ and $\mathbf a_{\mathrm{old}}$ denote actions from the aligned current and older views. For each mixed query $q$, $\mathbf r_{q,c}$ is the aligned reference consistent with camera $c$. Using all seven native action coordinates, we assign the query action to its closer reference:
\[
  \hat c_q=\operatorname*{arg\,min}_{c\in\{\mathrm{scene},\mathrm{wrist}\}}
  \|\mathbf a_q-\mathbf r_{q,c}\|_2,
  \qquad F_c=\frac{100}{2N}\sum_{q=1}^{2N}\mathbb{1}[\hat c_q=c].
\]
Each of the $N$ selected observations contributes both mixed queries with equal weight. In B/A, the older reference matches the scene camera; in A/B, it matches the wrist camera. No exact ties occur in the four-group summary. We retain observations whose aligned-reference separation is at or above the median for that policy and group.

Table~\ref{tab:temporal-view-following} lists the groups, counts, and sampling limits for Figure~\ref{fig:camera-conflict-temporal}. Early episode uses frames 10--19; near episode end uses the final ten demonstration frames, without alignment to first task completion. Groups are sampled separately and do not trace a common rollout. Near-end samples mostly come from opening, underrepresenting completions without opening.

For \pifive{}, scene-reference following is more frequent during transport (56.5--61.5\%), while wrist-reference following is more frequent in the other three groups. GR00T follows the wrist reference more often in all four groups, with transport closest to an even split.

\begin{table}[!htbp]
\centering
\small
\caption{Camera-reference following at a 10-frame lag, pooled equally over the B/A and A/B input pairs in Figure~\ref{fig:camera-conflict-temporal}. Each of $n$ selected observations contributes two queries. Shares use the closer full-seven-action reference; no exact ties occur. Within each policy and phase, selection retains reference separation at or above the median.}
\label{tab:temporal-view-following}
\begin{tabular}{lllrrr}
\toprule
Model & Proprio. & Phase & $n$ & Scene (\%) & Wrist (\%) \\
\midrule
\pifive{} & Yes & Early episode & 61 & 23.0 & 77.0 \\
\pifive{} & Yes & Near close & 100 & 41.5 & 58.5 \\
\pifive{} & Yes & Transport & 100 & 61.5 & 38.5 \\
\pifive{} & Yes & Near episode end & 65 & 27.7 & 72.3 \\
\pifive{} & No & Early episode & 61 & 30.3 & 69.7 \\
\pifive{} & No & Near close & 100 & 34.5 & 65.5 \\
\pifive{} & No & Transport & 100 & 56.5 & 43.5 \\
\pifive{} & No & Near episode end & 65 & 17.7 & 82.3 \\
GR00T & Yes & Early episode & 61 & 11.5 & 88.5 \\
GR00T & Yes & Near close & 100 & 21.0 & 79.0 \\
GR00T & Yes & Transport & 100 & 47.5 & 52.5 \\
GR00T & Yes & Near episode end & 65 & 30.0 & 70.0 \\
GR00T & No & Early episode & 61 & 24.6 & 75.4 \\
GR00T & No & Near close & 100 & 27.5 & 72.5 \\
GR00T & No & Transport & 100 & 43.5 & 56.5 \\
GR00T & No & Near episode end & 65 & 30.8 & 69.2 \\
\bottomrule
\end{tabular}
\par\smallskip
\begin{minipage}{.97\linewidth}
\footnotesize Early episode uses frames 10--19 (zero-based), the earliest ten frames permitting a 10-frame lag. Near close uses frames 4--12 before a sustained close command; transport uses the closed-command interval between closing and opening. Near episode end uses the final ten demonstration frames, not the first task-success time. For the early/end groups, 122/129 observations are available and 61/65 are selected per policy. Of the 129 available end observations, 128 were originally sampled around opening, so this group does not represent no-opening completions evenly. Groups use separately selected observations; connecting plot lines do not trace a shared rollout.
\end{minipage}
\end{table}

\FloatBarrier
\subsection{Spatial Translation-Action Projection}

Let $\mathbf u_{\mathrm{original}}$ and
$\mathbf u_{\mathrm{shifted}}$ denote the translation-action
references obtained when both views show the original and
shifted target positions, respectively.
Let $\mathbf u_c$ denote the translation action when only
camera $c$ shows the shifted target.
We compute
\[
\mathbf d =
\mathbf u_{\mathrm{shifted}}-\mathbf u_{\mathrm{original}},
\qquad
P_c =
\frac{
(\mathbf u_c-\mathbf u_{\mathrm{original}})^\top\mathbf d
}{
\|\mathbf d\|_2^2
}.
\]
A value of zero indicates no change along the reference
direction, while one indicates a projected change equal
to the reference shift.
Values can fall outside $[0,1]$, and the scene and wrist
scores need not sum to one.
We report the median score within each end-effector-to-target
distance bin: $[0,6)$, $[6,12)$, $[12,24)$, and $[24,40)$\,cm.

Table~\ref{tab:conflict-spatial-distance-lookup} lists medians and counts for Figure~\ref{fig:camera-conflict-spatial}. We retain probes with aligned translation-reference separation at or above the within-policy median, excluding those where either mixed seven-dimensional action equals A/A exactly. Both mixed conditions use these observations. Selection checks actions, not object visibility.

Wrist-view shifts produce larger median projections in every reported bin. Scene-view projections increase with end-effector distance, especially for \pifive{}. These target-shift responses may differ from camera influence under blackout or freezing.

\begin{table}[!htbp]
\centering
\small
\caption{Median XYZ action projection toward the displaced-target reference by end-effector distance. Distance bins are in cm and use $[\mathrm{lo},\mathrm{hi})$; $n$ is selected probes per bin and is shared by scene and wrist within a row.}
\label{tab:conflict-spatial-distance-lookup}
\begin{tabular}{llrrrr}
\toprule
Model & Proprio. & Distance (cm) & $n$ & Scene median & Wrist median \\
\midrule
\pifive{} & Yes & $[0,6)$ & 1119 & 0.030 & 0.874 \\
\pifive{} & Yes & $[6,12)$ & 569 & 0.115 & 0.772 \\
\pifive{} & Yes & $[12,24)$ & 386 & 0.268 & 0.706 \\
\pifive{} & Yes & $[24,40)$ & 166 & 0.377 & 0.613 \\
\pifive{} & No & $[0,6)$ & 1125 & 0.020 & 0.948 \\
\pifive{} & No & $[6,12)$ & 627 & 0.071 & 0.881 \\
\pifive{} & No & $[12,24)$ & 374 & 0.216 & 0.748 \\
\pifive{} & No & $[24,40)$ & 152 & 0.267 & 0.708 \\
GR00T & Yes & $[0,6)$ & 711 & 0.038 & 0.906 \\
GR00T & Yes & $[6,12)$ & 312 & 0.069 & 0.901 \\
GR00T & Yes & $[12,24)$ & 201 & 0.089 & 0.896 \\
GR00T & Yes & $[24,40)$ & 114 & 0.089 & 0.893 \\
GR00T & No & $[0,6)$ & 705 & 0.049 & 0.919 \\
GR00T & No & $[6,12)$ & 323 & 0.082 & 0.893 \\
GR00T & No & $[12,24)$ & 226 & 0.118 & 0.880 \\
GR00T & No & $[24,40)$ & 94 & 0.152 & 0.849 \\
\bottomrule
\end{tabular}
\end{table}

\FloatBarrier
\section{Additional Mitigation Results}
\label{sec:app-mitigation}

\subsection{Camera-Blackout Training Recipes and Suite Results}
\label{sec:app-camdrop-training}

\paragraph{Augmentation and optimization.}
Camera-specific training blacks out the selected view independently in 20\% of samples, leaving the other view intact. The lower-dose scene recipe uses 10\%. Either-camera training blacks out one uniformly chosen view in 20\% of samples: 10\% per camera, never both. All recipes retain the instruction, action target, and configured proprioception input.

We train the \pifive{} augmentation policies for 10,000 updates with batch size 32 and EMA decay 0.999, using the same learning-rate schedule as the control through the evaluated update. The principal runs use seed 43. The scene 10\%, scene 20\%, and either-camera recipes with proprioception also have seed-44 runs; reported ranges give the minimum and maximum across runs. GR00T uses the corresponding four-suite, 4,000-update recipe with one training seed per condition.

\paragraph{Suite results.}
Table~\ref{tab:mitigation-training-matrix} lists \pifive{} recipes and evaluation conditions; Table~\ref{tab:gr00t-camdrop-all-suite} reports GR00T results by suite (Spatial, Goal, and Object), with one training seed per recipe. Both use their own control cohorts, separate from the pooled baseline in Table~\ref{tab:trained-success}.

\begin{table}[t]
\centering
\small
\caption{Task success (\%) for the evaluated \pifive{} LIBERO-Spatial training recipe and proprioception condition. Entries are mean [minimum, maximum] across training runs; the dedicated $n_{\mathrm{runs}}$ column gives the number of runs.}
\label{tab:mitigation-training-matrix}
\begin{tabular}{lllrrr}
\toprule
Recipe & Proprio. & $n_{\mathrm{runs}}$ & Clean & Scene black & Scene frozen \\
\midrule
None & Yes & 1 & 93.3 & 22.0 & 2.7 \\
None & No & 1 & 93.3 & 14.0 & 1.3 \\
Scene 10\% & Yes & 2 & 96.0 [95.3,96.7] & 91.0 [89.3,92.7] & 34.3 [30.0,38.7] \\
Scene 10\% & No & 1 & 92.7 & 86.0 & 14.7 \\
Scene 20\% & Yes & 2 & 94.7 [94.0,95.3] & 94.0 [93.3,94.7] & 53.0 [45.3,60.7] \\
Scene 20\% & No & 1 & 95.3 & 91.3 & 38.0 \\
Wrist 20\% & Yes & 1 & 93.3 & 16.7 & 0.0 \\
Wrist 20\% & No & 1 & 93.3 & 3.3 & 0.0 \\
Either 10\%/view & Yes & 2 & 94.3 [92.7,96.0] & 85.7 [84.0,87.3] & 12.7 [10.0,15.3] \\
Either 10\%/view & No & 1 & 90.7 & 84.7 & 2.0 \\
\bottomrule
\end{tabular}
\medskip
\begin{tabular}{lllrr}
\toprule
Recipe & Proprio. & $n_{\mathrm{runs}}$ & Wrist black & Wrist frozen \\
\midrule
None & Yes & 1 & 20.7 & 0.0 \\
None & No & 1 & 0.7 & 0.0 \\
Scene 10\% & Yes & 2 & 18.3 [12.7,24.0] & 0.0 [0.0,0.0] \\
Scene 10\% & No & 1 & 0.7 & 0.0 \\
Scene 20\% & Yes & 2 & 14.3 [10.7,18.0] & 0.0 [0.0,0.0] \\
Scene 20\% & No & 1 & 1.3 & 0.0 \\
Wrist 20\% & Yes & 1 & 73.3 & 30.0 \\
Wrist 20\% & No & 1 & 52.7 & 0.0 \\
Either 10\%/view & Yes & 2 & 68.3 [65.3,71.3] & 8.7 [6.0,11.3] \\
Either 10\%/view & No & 1 & 51.3 & 0.0 \\
\bottomrule
\end{tabular}
\end{table}

\begin{table*}[t]
\centering
\small
\setlength{\tabcolsep}{3.5pt}
\begin{tabular}{lllccccc}
\toprule
Suite & Proprio. & Training & Clean & \shortstack{Scene\\black} & \shortstack{Scene\\frozen} & \shortstack{Wrist\\black} & \shortstack{Wrist\\frozen} \\
\midrule
Spatial & Yes & Plain & 99.3 & 44.7 & 7.3 & 8.0 & 0.0 \\
Spatial & Yes & Scene blackout & 98.7 & 99.3 & 58.0 & 7.3 & 0.0 \\
Spatial & Yes & Wrist blackout & 98.7 & 34.7 & 0.0 & 93.3 & 0.0 \\
Spatial & Yes & Scene + wrist & 100.0 & 96.0 & 7.3 & 88.0 & 0.0 \\
\addlinespace
Spatial & No & Plain & 97.3 & 32.7 & 0.7 & 0.0 & 0.0 \\
Spatial & No & Scene blackout & 99.3 & 97.3 & 6.7 & 0.0 & 0.0 \\
Spatial & No & Wrist blackout & 99.3 & 35.3 & 0.0 & 79.3 & 0.0 \\
Spatial & No & Scene + wrist & 93.3 & 95.3 & 0.7 & 72.0 & 0.0 \\
\addlinespace
\midrule
Goal & Yes & Plain & 97.3 & 52.7 & 20.7 & 14.0 & 0.0 \\
Goal & Yes & Scene blackout & 97.3 & 96.7 & 74.7 & 8.7 & 0.0 \\
Goal & Yes & Wrist blackout & 97.3 & 48.0 & 3.3 & 93.3 & 6.0 \\
Goal & Yes & Scene + wrist & 97.3 & 94.0 & 15.3 & 90.0 & 0.0 \\
\addlinespace
Goal & No & Plain & 94.7 & 57.3 & 14.7 & 0.0 & 0.0 \\
Goal & No & Scene blackout & 97.3 & 90.0 & 47.3 & 0.7 & 0.0 \\
Goal & No & Wrist blackout & 98.0 & 39.3 & 0.0 & 88.7 & 0.0 \\
Goal & No & Scene + wrist & 98.7 & 90.7 & 10.7 & 75.3 & 0.0 \\
\addlinespace
\midrule
Object & Yes & Plain & 99.3 & 57.3 & 0.0 & 14.7 & 0.0 \\
Object & Yes & Scene blackout & 100.0 & 99.3 & 75.3 & 8.7 & 0.0 \\
Object & Yes & Wrist blackout & 99.3 & 44.0 & 0.0 & 96.7 & 0.7 \\
Object & Yes & Scene + wrist & 98.7 & 99.3 & 0.7 & 92.0 & 0.0 \\
\addlinespace
Object & No & Plain & 99.3 & 13.3 & 0.0 & 0.0 & 0.0 \\
Object & No & Scene blackout & 100.0 & 100.0 & 32.7 & 0.0 & 0.0 \\
Object & No & Wrist blackout & 100.0 & 7.3 & 0.0 & 90.0 & 0.0 \\
Object & No & Scene + wrist & 99.3 & 100.0 & 0.0 & 87.3 & 0.0 \\
\addlinespace
\bottomrule
\end{tabular}
\caption{GR00T task success at episode-start faults across Spatial, Goal, and Object (\%; 150 rollouts per cell). Yes and No indicate policies with and without proprioception. Training uses 20\% camera blackout. One training seed per recipe.}
\label{tab:gr00t-camdrop-all-suite}
\end{table*}

\FloatBarrier
\subsection{Component Transplants after Blackout Training}
\label{sec:app-component-transplants}

We transfer the vision encoder and projector, VLM backbone, or action expert from the scene-blackout-trained \pifive{} policy with proprioception into its unaugmented counterpart. Each transplant leaves other weights unchanged. Evaluation uses 150 LIBERO-Spatial episodes per condition, with scene blackout from episode start.

The VLM transplant raises success from 22.7\% to 90.0\%, versus 38.0\% for the vision encoder and projector and 26.0\% for the action expert (Table~\ref{tab:training-component-transplants}). This recovers most of the gain of the fully trained donor, which reaches 94.7\% in a separate evaluation, without identifying the responsible changes within the VLM.

\begin{table}[!htbp]
\centering
\caption{Component transplants from a scene-blackout-trained \pifive{} policy with proprioception into the corresponding policy trained without blackout.}
\label{tab:training-component-transplants}

\resizebox{0.4\linewidth}{!}{%
\begin{tabular}{@{}lr@{}}
\toprule
Transplanted component & Success (\%) \\
\midrule
None & 22.7 \\
Vision encoder + projector & 38.0 \\
VLM backbone & \textbf{90.0} \\
Action expert & 26.0 \\
\midrule
All components (trained policy) & 94.7 \\
\bottomrule
\end{tabular}%
}
\end{table}

\FloatBarrier
\subsection{Physical Outcomes after Blackout Training}
\begin{figure}[!htbp]
  \centering
  \includegraphics[width=.92\linewidth]{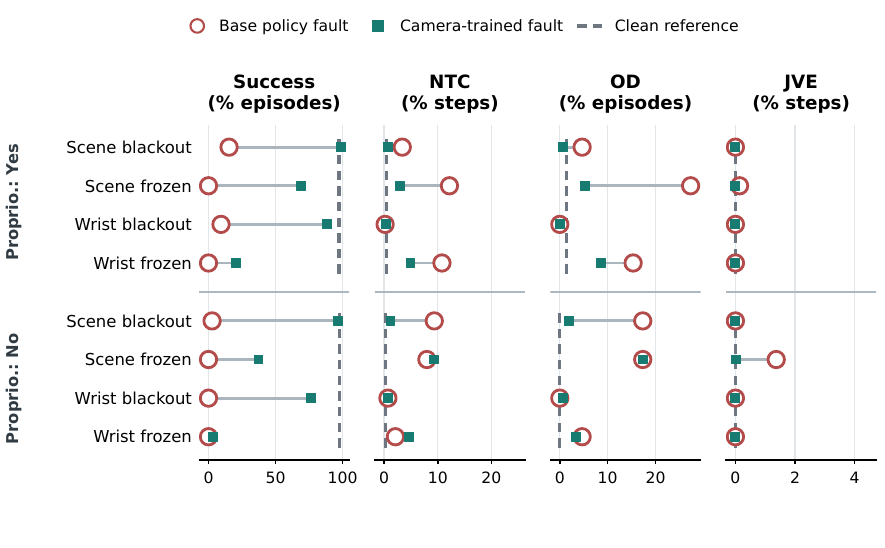}
  \caption{\pifive{} task success and physical indicators after blackout training on Object. Open circles, squares, and dashed lines denote base-fault, camera-trained-fault, and base-clean results. Physical indicators use the first 105 executed steps. \texttt{NTC} and \texttt{JVE} report step percentages; \texttt{OD} reports episode percentages.}
  \label{fig:physical-recovery-object}
\end{figure}

\begin{figure}[!htbp]
  \centering
  \includegraphics[width=.92\linewidth]{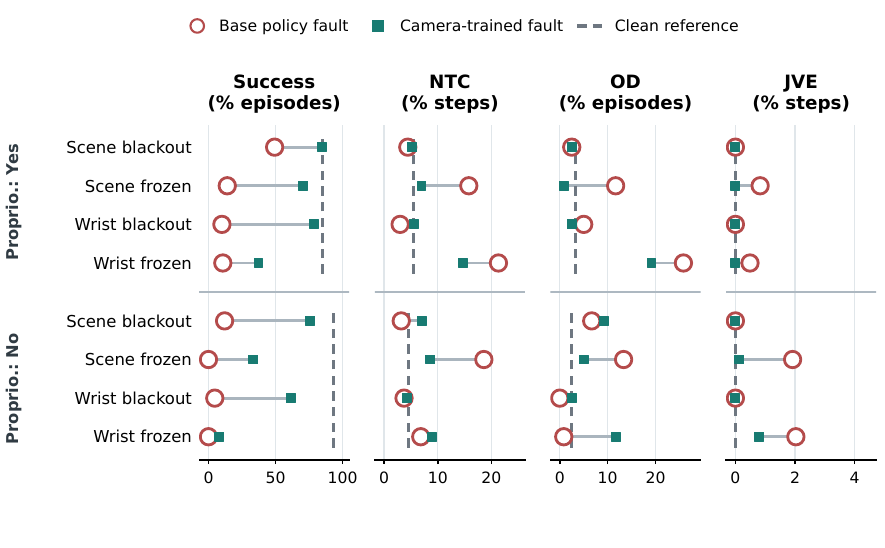}
  \caption{\pifive{} task success and physical indicators after blackout training on Goal. Open circles, squares, and dashed lines denote base-fault, camera-trained-fault, and base-clean results. Physical indicators use the first 105 executed steps. \texttt{NTC} and \texttt{JVE} report step percentages; \texttt{OD} reports episode percentages. \texttt{NTC} and \texttt{OD} use the 120 episodes with a defined target out of 150 rollouts per condition.}
  \label{fig:physical-recovery-goal}
\end{figure}

Table~\ref{tab:physical-recovery-spatial} provides the Spatial values accompanying Figure~\ref{fig:physical-recovery}. Figures~\ref{fig:physical-recovery-object} and~\ref{fig:physical-recovery-goal} extend the same measurements to Object and Goal. We record success over the whole episode and physical indicators over up to 105 executed steps. \texttt{NTC} and \texttt{JVE} report step percentages; \texttt{OD} reports episode percentages. For Goal, non-target metrics use only the 120 rollouts with a defined target.

\begin{table}[!htbp]
\caption{Task success and physical outcomes after blackout training on LIBERO-Spatial. SR is whole-episode success; contact and joint-speed exceedance are step rates, and displacement reports sustained movement of a non-target object. Physical measurements use up to 105 steps.}
\label{tab:physical-recovery-spatial}
\centering
\scriptsize
\setlength{\tabcolsep}{2pt}
\begin{tabular}{llllrrrr}
\toprule Proprio. & Camera & Fault & Policy/input & SR & NTC & JVE & OD \\
\midrule
Yes & scene & black & Control / clean & 0.953 & 0.025 & 0.000 & 0.013 \\
Yes & scene & black & Control / fault & 0.233 & 0.054 & 0.000 & 0.053 \\
Yes & scene & black & Trained / fault & 0.953 & 0.021 & 0.000 & 0.027 \\
\addlinespace
Yes & scene & frozen & Control / clean & 0.953 & 0.025 & 0.000 & 0.013 \\
Yes & scene & frozen & Control / fault & 0.027 & 0.101 & 0.018 & 0.253 \\
Yes & scene & frozen & Trained / fault & 0.587 & 0.039 & 0.000 & 0.060 \\
\addlinespace
Yes & wrist & black & Control / clean & 0.953 & 0.025 & 0.000 & 0.013 \\
Yes & wrist & black & Control / fault & 0.213 & 0.012 & 0.000 & 0.020 \\
Yes & wrist & black & Trained / fault & 0.747 & 0.022 & 0.000 & 0.013 \\
\addlinespace
Yes & wrist & frozen & Control / clean & 0.953 & 0.025 & 0.000 & 0.013 \\
Yes & wrist & frozen & Control / fault & 0.000 & 0.120 & 0.004 & 0.233 \\
Yes & wrist & frozen & Trained / fault & 0.287 & 0.095 & 0.000 & 0.093 \\
\addlinespace
No & scene & black & Control / clean & 0.947 & 0.023 & 0.000 & 0.027 \\
No & scene & black & Control / fault & 0.127 & 0.021 & 0.000 & 0.053 \\
No & scene & black & Trained / fault & 0.907 & 0.023 & 0.000 & 0.013 \\
\addlinespace
No & scene & frozen & Control / clean & 0.947 & 0.023 & 0.000 & 0.027 \\
No & scene & frozen & Control / fault & 0.013 & 0.073 & 0.042 & 0.227 \\
No & scene & frozen & Trained / fault & 0.387 & 0.047 & 0.006 & 0.093 \\
\addlinespace
No & wrist & black & Control / clean & 0.947 & 0.023 & 0.000 & 0.027 \\
No & wrist & black & Control / fault & 0.000 & 0.004 & 0.002 & 0.007 \\
No & wrist & black & Trained / fault & 0.560 & 0.035 & 0.000 & 0.013 \\
\addlinespace
No & wrist & frozen & Control / clean & 0.947 & 0.023 & 0.000 & 0.027 \\
No & wrist & frozen & Control / fault & 0.007 & 0.059 & 0.029 & 0.113 \\
No & wrist & frozen & Trained / fault & 0.000 & 0.100 & 0.004 & 0.093 \\
\bottomrule
\end{tabular}
\end{table}

\FloatBarrier
\subsection{Mean-Embedding Replacement: Calibration and Fault-Onset Results}
\label{sec:app-embedding-replacement}

\paragraph{Calibration and replacement.}
For camera $c$ and visual-token position $p$, let $h_{c,p}(I_i)$ be the representation of a clean calibration image. We compute
\begin{equation}
\bar h_{c,p}=\frac{1}{N}\sum_{i=1}^{N}h_{c,p}(I_i),
\qquad
h_{c,p}(I_{\mathrm{fault}})\leftarrow\bar h_{c,p}.
\label{eq:appendix-mean-embedding}
\end{equation}
We average separately by camera and token position, with policy weights fixed. The mean of clean-image representations need not correspond to a realizable image.

For \pifive{}, we take 2,550 frames from the Spatial training data at stride 20. Every tenth episode is reserved for separate open-loop screening and excluded here. We compute means separately for the policies with and without proprioception, then insert them after vision encoding and projection, at the first VLM layer's input.

GR00T uses clean rollouts from six held-out initial states per Spatial task (indices 15--20), separate from evaluation indices 0--14. The mean uses 830 observation requests with proprioception and 799 without it. We replace both the primary image embeddings and the additional visual features injected into the language backbone; otherwise, those features would still carry the affected camera's information. This calibration source and replacement interface differ from \pifive{}'s, which uses 2,550 training frames.

\paragraph{Trigger.}
Replacement starts when all pixels are zero or the image exactly repeats the one from the preceding policy request. We clear detector history at each episode, so the first frozen frame may pass through before repetition is detected. Oracle variants, reported separately, start at the known fault time. The other camera's tokens remain available, and calibration uses no fault-evaluation outcomes.

\begin{figure}[!htbp]
  \centering
  \includegraphics[width=\linewidth]{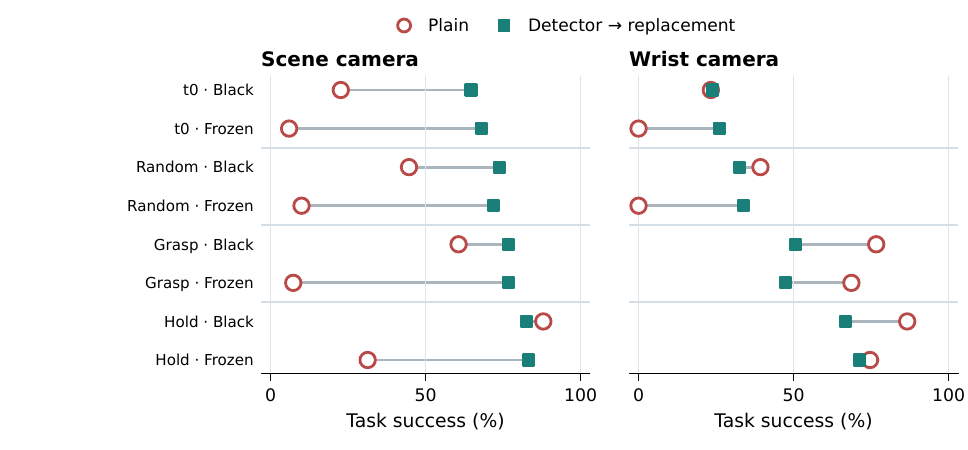}
  \caption{\pifive{} task success with proprioception under detector-triggered mean-embedding replacement. Open circles denote the unmodified policy and squares the replacement intervention. Each comparison uses 150 Spatial episodes paired within one evaluation setup.}
  \label{fig:app-trainfree-embedding-onsets}
\end{figure}

\paragraph{Results across fault onsets.}
Figure~\ref{fig:app-trainfree-embedding-onsets} covers random, grasp, and hold fault onsets. Paired success differences are listed in Table~\ref{tab:trainfree-detector-safety}, with bootstrap intervals resampled by task. For \texttt{NTC} and \texttt{OD}, Table~\ref{tab:trainfree-detector-safety-physical} uses up to 105 executed steps and resamples episodes for the intervals.

\begin{table}[t]
\centering\small
\caption{Task success without and with detector-triggered mean-embedding replacement for camera faults from episode start and random, grasp, and hold onsets. Start denotes episode start. Each paired LIBERO-Spatial condition has 150 episodes. Differences are absolute changes in success (\%) with 95\% task-cluster bootstrap intervals; $p$ is exact McNemar.}
\label{tab:trainfree-detector-safety}
\begin{tabular}{lrrrr}
\toprule
Fault & Original & Replacement & $\Delta$ success (\%) [95\%] & McNemar $p$ \\
\midrule
Scene start · Black & 34/150 & 97/150 & +42.0 [+26.7, +58.7] & 3.4e-15 \\
Scene start · Frozen & 9/150 & 102/150 & +62.0 [+50.7, +73.3] & 2.0e-28 \\
Wrist start · Black & 35/150 & 36/150 & +0.7 [-10.0, +11.3] & 1.000 \\
Wrist start · Frozen & 0/150 & 39/150 & +26.0 [+14.7, +38.0] & 3.6e-12 \\
Scene Random · Black & 67/150 & 111/150 & +29.3 [+5.3, +52.7] & 1.7e-07 \\
Scene Random · Frozen & 15/150 & 108/150 & +62.0 [+48.0, +74.0] & 6.0e-26 \\
Wrist Random · Black & 59/150 & 49/150 & -6.7 [-19.3, +7.3] & 0.203 \\
Wrist Random · Frozen & 0/150 & 51/150 & +34.0 [+20.7, +48.0] & 8.9e-16 \\

Scene Grasp · Black & 91/150 & 115/150 & +16.0 [-4.0, +36.0] & 0.00369 \\
Scene Grasp · Frozen & 11/150 & 115/150 & +69.3 [+56.7, +81.3] & 3.42e-28 \\
Wrist Grasp · Black & 115/150 & 76/150 & -26.0 [-36.0, -14.0] & 4e-08 \\
Wrist Grasp · Frozen & 103/150 & 71/150 & -21.3 [-38.7, -2.0] & 0.000166 \\
Scene Hold · Black & 132/150 & 124/150 & -5.3 [-17.3, +6.0] & 0.169 \\
Scene Hold · Frozen & 47/150 & 125/150 & +52.0 [+32.0, +71.3] & 6.62e-24 \\
Wrist Hold · Black & 130/150 & 100/150 & -20.0 [-32.7, -9.3] & 2.83e-06 \\
Wrist Hold · Frozen & 112/150 & 107/150 & -3.3 [-25.3, +21.3] & 0.551 \\
\bottomrule
\end{tabular}
\end{table}

\begin{table}[t]
\centering\small
\caption{Changes in physical outcomes after detector-triggered mean-embedding replacement for the same paired conditions as Table~\ref{tab:trainfree-detector-safety}. Each cell reports replacement minus original with an episode-level 95\% bootstrap interval, using up to 105 executed steps. Differences are absolute changes in NTC (\% of steps) and OD (\% of episodes). OD records non-target displacement above 2\,cm for six consecutive samples.}
\label{tab:trainfree-detector-safety-physical}
\begin{tabular*}{\linewidth}{@{}@{\extracolsep{\fill}}lrr@{}}
\toprule
Fault & $\Delta$ NTC (\%) [95\% CI] & $\Delta$ OD (\%) [95\% CI] \\
\midrule
Scene start · Black & -0.5 [-2.6, +1.8] & +6.7 [+0.7, +13.3] \\
Scene start · Frozen & -5.0 [-7.5, -2.5] & -15.3 [-23.3, -7.3] \\
Wrist start · Black & +2.1 [+0.9, +3.5] & +2.0 [+0.0, +4.7] \\
Wrist start · Frozen & -8.6 [-11.3, -6.1] & -16.0 [-22.0, -10.0] \\
Scene Random · Black & +0.1 [-0.9, +1.2] & +2.0 [-2.0, +6.7] \\
Scene Random · Frozen & -4.9 [-7.1, -2.7] & -17.3 [-24.0, -10.7] \\
Wrist Random · Black & +1.5 [+0.8, +2.3] & +0.0 [-2.7, +2.7] \\
Wrist Random · Frozen & -13.1 [-16.4, -9.8] & -16.7 [-23.3, -10.0] \\
Scene Grasp · Black & +0.1 [-0.5, +0.6] & +1.3 [-1.3, +4.0] \\
Scene Grasp · Frozen & +0.7 [+0.2, +1.1] & +0.7 [-1.3, +3.3] \\
Wrist Grasp · Black & +0.3 [+0.1, +0.5] & +0.7 [-1.3, +3.3] \\
Wrist Grasp · Frozen & +0.1 [-0.1, +0.4] & +0.0 [+0.0, +0.0] \\
Scene Hold · Black & +0.1 [-0.1, +0.3] & +0.0 [+0.0, +0.0] \\
Scene Hold · Frozen & +0.5 [+0.2, +0.8] & +0.0 [-2.0, +2.0] \\
Wrist Hold · Black & +0.1 [-0.0, +0.2] & +0.0 [+0.0, +0.0] \\
Wrist Hold · Frozen & -0.1 [-0.4, +0.0] & -1.3 [-3.3, +0.0] \\
\bottomrule
\end{tabular*}
\end{table}

\FloatBarrier
\subsection{Attention Changes after Blackout Training}
\label{sec:app-attention-mitigation}

\begin{figure*}[!htbp]
  \centering
  \includegraphics[width=.94\textwidth]{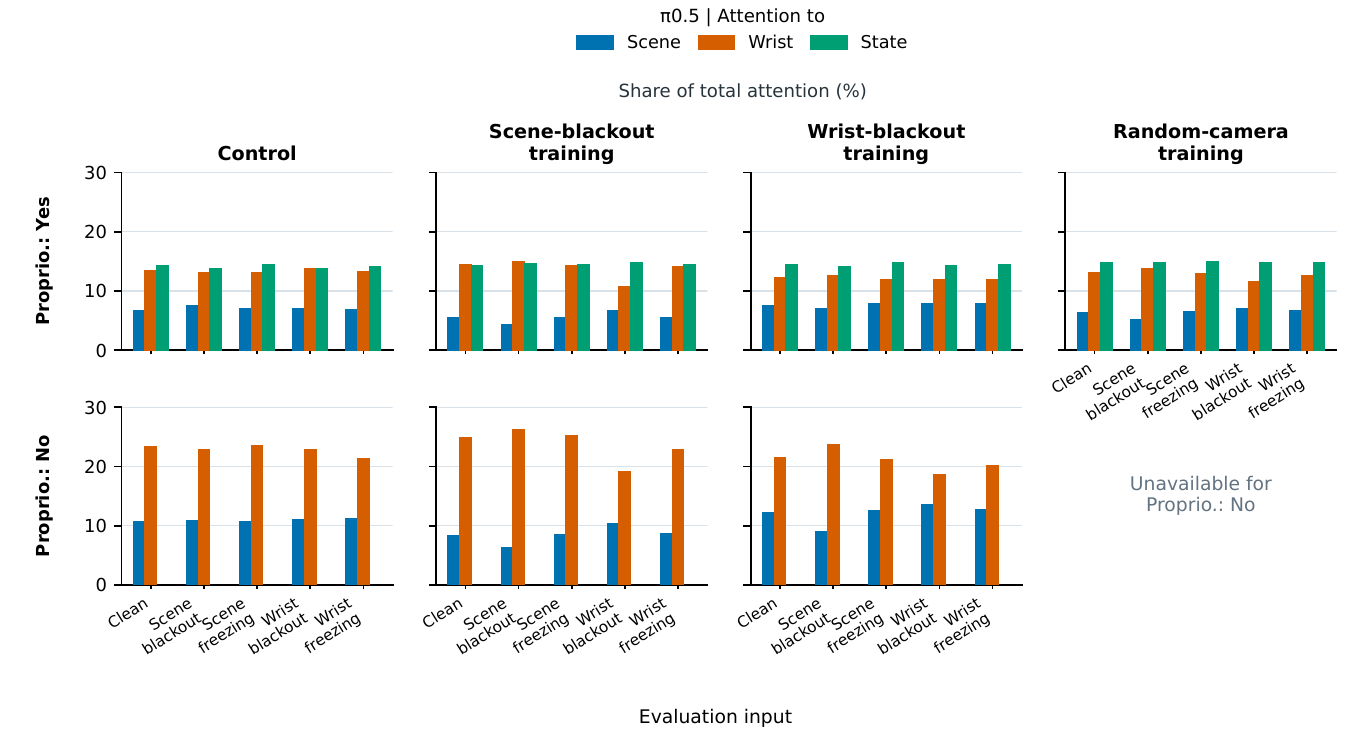}
  \caption{Absolute input-group attention masses for scene vision, wrist vision, and state across clean and camera-fault probes. Values are attention masses multiplied by 100, measured over 96 fixed-probe episodes across four suites. This probe cohort is separate from the LIBERO-Spatial gain rollouts.}
  \label{fig:app-attention-training-input-mass}
\end{figure*}

Figure~\ref{fig:app-attention-training-input-mass} plots attention shares for control and trained policies on fixed probes. These probes use a separate cohort from the success rollouts and cannot establish that attention changes cause recovery. Appendix~\ref{sec:additional-attention-reweighting} tests inference-time attention reweighting.

\FloatBarrier
\raggedbottom
\section{Additional Mitigation Attempts}
\label{sec:additional-mitigation-attempts}

We test attention reweighting, camera-token masking, and older-observation training in \pifive{} on LIBERO-Spatial. Each experiment has its own control checkpoint and evaluation cohort. Results cover task success under clean inputs, blackout, and freezing at the specified budgets and intervention strengths; physical safety is not evaluated.

\FloatBarrier
\subsection{Inference-Time Attention Reweighting}
\label{sec:additional-attention-reweighting}

\paragraph{Intervention.}
We add bias $g$ to a token group's pre-softmax attention logits, multiplying its unnormalized weights by $e^g$; $g=0$ leaves them unchanged. Groups comprise state, scene, and wrist tokens. Biases apply in all 18 prefix-transformer layers and to action queries in all 18 action-expert layers, so they can change both prefix representations and the expert's attention to them. We assume the faulty camera is known and test the same biases on clean inputs.

\paragraph{Increasing attention to proprioception does not recover performance.}
We test manually selected biases in the policy with proprioception, trained for 10,000 updates with EMA. Each condition uses 150 episodes: 15 initial states for each of ten tasks. Increasing the state-token bias under scene blackout lowers success from 24.7\% at zero bias to 0\% at $g_{\mathrm{state}}=4$ (Table~\ref{tab:additional-attention-manual}). Suppressing scene tokens alone changes success much less. A positive state bias also lowers clean-input success.

\begin{table}[!htbp]
\caption{Task success under manually selected attention biases. Each row uses 150 Spatial episodes and the same \pifive{} checkpoint with proprioception. Unlisted token groups have zero bias.}
\label{tab:additional-attention-manual}
\centering\small
\setlength{\tabcolsep}{8pt}
\begin{tabular*}{\linewidth}{@{}@{\extracolsep{\fill}}lrrr@{}}
\toprule
Input & State bias & Scene bias & Success (\%) \\
\midrule
Scene blackout & 0 & 0 & 24.7 \\
 & 0.25 & 0 & 23.3 \\
 & 0.5 & 0 & 21.3 \\
 & 1 & 0 & 14.7 \\
 & 2 & 0 & 4.0 \\
 & 4 & 0 & 0.0 \\
\cmidrule(lr){2-4}
 & 0 & $-1$ & 26.0 \\
 & 0 & $-2$ & 23.3 \\
 & 1 & $-1$ & 18.0 \\
 & 2 & $-2$ & 5.3 \\
\midrule
Clean & 0 & 0 & 96.0 \\
 & 2 & 0 & 43.3 \\
 & 2 & $-2$ & 20.7 \\
\bottomrule
\end{tabular*}
\end{table}

\paragraph{Matching selected attention ratios does not reproduce the trained policy's robustness.}
We calibrate biases on fixed observation probes to match the state, scene, and wrist attention ratios of camera-blackout-trained policies, then evaluate on a separate Spatial cohort. The group ratios match within 1\% relative error, but success recovers little of the trained-policy gain (Table~\ref{tab:additional-attention-matched}). Under scene blackout, the full adjustment recovers 6.4\% of the gain, with a 95\% interval of [0.0, 12.9]\%. No success improvement survives Holm correction across the eight tested settings. Matching selected group ratios alone does not test the full trained attention distribution or rule out its role in robustness.

\begin{table}[!htbp]
\caption{Attention biases calibrated to blackout-trained policies. Success uses 150 episodes per cell. Full adjustment targets state and camera groups; camera-only adjustment targets camera groups. The trained reference is a separate policy trained with blackout in the evaluated camera. A dash denotes a setting not shown in this summary.}
\label{tab:additional-attention-matched}
\centering\small
\begin{tabular*}{\linewidth}{@{}@{\extracolsep{\fill}}lcc@{}}
\toprule
Policy or intervention & Scene blackout (\%) & Wrist blackout (\%) \\
\midrule
Unmodified control & 22.0 & 21.3 \\
Full group adjustment & 26.7 & 22.7 \\
Camera-only adjustment & --- & 26.7 \\
Blackout-trained reference & 94.7 & 72.7 \\
\bottomrule
\end{tabular*}
\end{table}

\begin{figure}[!htbp]
  \centering
  \includegraphics[width=\linewidth]{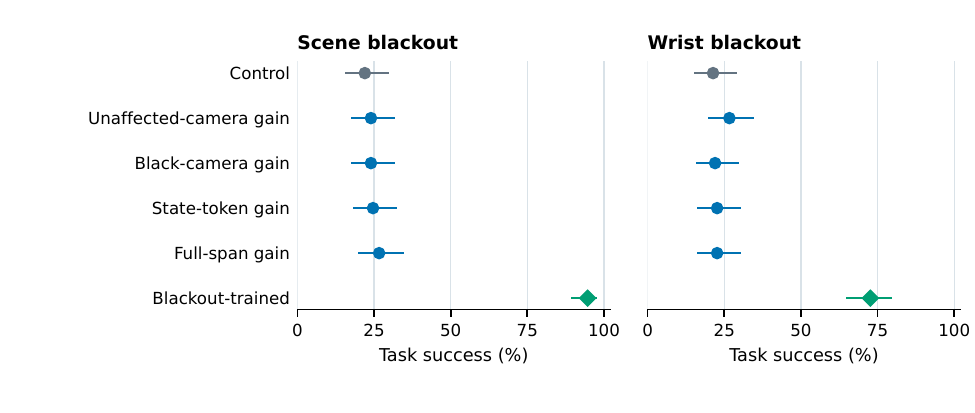}
  \caption{Inference-time attention reweighting under matched scene- and wrist-blackout conditions. Each arm contains 150 Spatial episodes; intervals are 95\% Wilson confidence intervals. Trained-checkpoint controls remain distinct from inference-only interventions.}
  \label{fig:attention-gain-outcomes}
\end{figure}

Figure~\ref{fig:attention-gain-outcomes} compares individual and full token-group adjustments with blackout-trained references. Appendix~\ref{sec:app-attention-mitigation} reports attention changes after training, separately from these inference-only interventions.

\FloatBarrier
\subsection{Oracle Camera-Token Masking}
\label{sec:additional-oracle-masking}

\paragraph{Intervention and scope.}
With the faulty camera known, expert-only masking blocks its tokens as action-expert keys and values but allows them to affect other tokens in the prefix transformer. Prefix-and-expert masking blocks access in both modules. Faults and masks start at episode start; freezing repeats the initial frame. We test the released LIBERO checkpoint and separate 10,000-update checkpoints trained with and without proprioception, without EMA.

\paragraph{The outcome depends on both checkpoint and masking scope.}
Expert-only masking leaves wrist-freezing success at zero in all three policies (Table~\ref{tab:additional-oracle-masking}). Prefix-and-expert masking raises scene-freezing success from 1.3\% to 47.3\% for the released checkpoint, with no comparable gain in the fine-tuned pair. Recovery from faults arising during manipulation remains untested.

\begin{table}[!htbp]
\caption{Task success (\%) under oracle camera-token masking, with 150 episodes per cell. The two fine-tuned policies are a separate non-EMA checkpoint pair. All faults and masks start at episode start. Dashes indicate unevaluated settings.}
\label{tab:additional-oracle-masking}
\centering\small
\setlength{\tabcolsep}{5pt}
\begin{tabular*}{\linewidth}{@{}@{\extracolsep{\fill}}llccc@{}}
\toprule
Policy & Camera fault & No mask & Expert only & Prefix + expert \\
\midrule
Released & Scene freezing & 1.3 & 4.0 & 47.3 \\
 & Wrist freezing & 0.0 & 0.0 & --- \\
\midrule
Fine-tuned, with proprio. & Scene freezing & 25.3 & 30.0 & 18.0 \\
 & Wrist freezing & 0.0 & 0.0 & --- \\
\midrule
Fine-tuned, without proprio. & Scene freezing & 6.7 & 14.0 & 10.7 \\
 & Wrist freezing & 0.0 & 0.0 & --- \\
\bottomrule
\end{tabular*}
\end{table}

\FloatBarrier
\subsection{Training with Older Observations}
\label{sec:additional-older-observation-training}

\paragraph{Optimization.}
Training uses batch size 32, seed 42, and AdamW with $(\beta_1,\beta_2)=(0.9,0.95)$, $\epsilon=10^{-8}$, weight decay $10^{-10}$, and gradient-norm clipping at 1.0. The learning rate warms up linearly over 10,000 updates to $5\times10^{-5}$, then stays constant; 4,000-update runs end during warmup. Evaluation uses unaveraged weights, unlike the principal simulation results that use EMA. Each recipe, including the separate 12,000-update freshness experiment, has one training run per condition.

\paragraph{Short-budget training and camera-freshness indicators.}
We train the base \pifive{} checkpoint for 4,000 updates. In 10\% of samples, one uniformly chosen camera receives an image from 4--32 frames earlier, clipped to available history; the action target stays current. We compare augmentation alone with augmentation plus two binary freshness indicators (current: one; replaced: zero). A zero-initialized state projection receives the indicators with robot-state coordinates zeroed. The comparison therefore adds both freshness information and an input pathway, without proprioception.

Augmentation alone raises scene-freezing success from 4\% to 22\% and leaves wrist-freezing success at 2\% (Table~\ref{tab:additional-freshness-training}; 50 episodes per condition). Adding freshness indicators raises wrist-freezing success to 10\%, though the difference is not significant (two-sided exact McNemar $p=0.219$). Forcing the indicators to report current images removes all five successes ($p=0.0625$). At 12,000 updates, freshness indicators change success by $+6\%$ absolute under wrist freezing ($p=0.508$) and $-8\%$ absolute under scene freezing ($p=0.523$), relative to augmentation alone. Neither comparison is significant. These longer runs have no unaugmented control.

\begin{table}[!htbp]
\caption{Task success (\%) after 4,000 updates from the base checkpoint, with 50 episodes per cell. Freezing begins after three consecutive gripper-close commands. All scheduled episodes remain in the denominator, including those that do not reach the closing trigger. The control receives no temporal augmentation.}
\label{tab:additional-freshness-training}
\centering\small
\begin{tabular*}{\linewidth}{@{}@{\extracolsep{\fill}}lccc@{}}
\toprule
Evaluation input & Control & Augmentation & Augmentation + freshness \\
\midrule
Clean & 42.0 & 36.0 & 46.0 \\
Scene freezing & 4.0 & 22.0 & 12.0 \\
Wrist freezing & 0.0 & 2.0 & 10.0 \\
\bottomrule
\end{tabular*}
\end{table}

\FloatBarrier
\paragraph{Continued training from a competent LIBERO policy.}
We continue training the released LIBERO policy and a matched unaugmented control for 4,000 updates. In 10\% of samples, augmentation attempts to replace one uniformly chosen camera image. The replacement is equally likely to come from the demonstration's first sustained closing landmark or from 8--40 frames earlier. A second recipe shortens the delay to 2--8 frames.

We skip ineligible pairs and pairs with mean absolute pixel difference below about 8.36 on the 0--255 scale, without drawing a replacement pair. The threshold is the 25th percentile of 2,000 calibration pairs from the longer-delay recipe. After filtering, replacement rates are 5.7\% and 5.0\%, so the recipes differ in exposure as well as image age. A third variant adds vector-field consistency with weight one. It uses the current-observation prediction as a stop-gradient target, with the same noise and diffusion time for both inputs.

Table~\ref{tab:additional-continued-training} shows limited recovery under freezing. The 8--40-frame mixture preserves 98.7\% clean success and raises scene-freezing success from 0\% to 7.3\%; the 2--8-frame mixture reaches 1.3\%. Both leave wrist-freezing and wrist-blackout success at zero. Adding consistency lowers clean success to 73.3\%, with 4.7\% scene-freezing success and no wrist-freezing recovery.

\begin{table}[!htbp]
\caption{Task success (\%) after 4,000 additional updates from the released LIBERO policy, with 150 episodes per cell. Both temporal mixtures contain landmark freezing and delayed images; only the sampled delay range differs. Consistency uses the 8--40-frame mixture. Faults begin at episode start.}
\label{tab:additional-continued-training}
\centering\small
\setlength{\tabcolsep}{5pt}
\begin{tabular*}{\linewidth}{@{}@{\extracolsep{\fill}}lcccc@{}}
\toprule
 & & \multicolumn{2}{c}{Temporal mixture} & \\
\cmidrule(lr){3-4}
Evaluation input & Control & Delay 8--40 & Delay 2--8 & + Consistency \\
\midrule
Clean & 98.7 & 98.7 & 98.7 & 73.3 \\
Scene freezing & 0.0 & 7.3 & 1.3 & 4.7 \\
Wrist freezing & 0.0 & 0.0 & 0.0 & 0.0 \\
Wrist blackout & 0.0 & 0.0 & 0.0 & 0.0 \\
\bottomrule
\end{tabular*}
\end{table}

\FloatBarrier
\end{document}